\documentclass[sn-mathphys-num]{sn-jnl}

\usepackage{graphicx}%
\usepackage{multirow}%
\usepackage{amsmath,amssymb,amsfonts}%
\usepackage{amsthm}%
\usepackage{mathrsfs}%
\usepackage[title]{appendix}%
\usepackage{xcolor}%
\usepackage{textcomp}%
\usepackage{manyfoot}%
\usepackage{booktabs}%
\usepackage{algorithm}%
\usepackage{algorithmicx}%
\usepackage{algpseudocode}%
\usepackage{listings}%
\usepackage{amsmath,amssymb,amsfonts}
\usepackage{graphicx}
\usepackage{textcomp}
\usepackage{xcolor}
\usepackage{pbox}
\usepackage{multirow}
\usepackage{lipsum}
\usepackage{tabulary}
\usepackage{array}
\usepackage{hhline}
\usepackage{subcaption}
\usepackage{textcomp,mathcomp}
\usepackage{gensymb}
\usepackage{rotating}
\usepackage{array,multirow}
\usepackage{pdflscape}
\usepackage{afterpage}
\usepackage{float}
\usepackage{algorithm}
\usepackage{xurl}
\usepackage[algo2e, norelsize, linesnumbered, ruled, lined, boxed, commentsnumbered]{algorithm2e}

\theoremstyle{thmstyleone}%
\theoremstyle{thmstyletwo}%

\theoremstyle{thmstylethree}%
\usepackage{lmodern}

\usepackage{fancyhdr} 

\begin{document}

\title[Article Title]{Article Title}

\title[Article Title]{Efficient Deterministic Renewable Energy Forecasting Guided by Multiple-Location Weather Data 
}

\author*[1]{\fnm{Charalampos} \sur{Symeonidis}}

\author[2]{\fnm{Nikos} \sur{Nikolaidis}}

\affil[1,2]{\orgdiv{Department of Informatics}, \orgname{Aristotle University of Thessaloniki}, \orgaddress{ \city{Thessaloniki}, \postcode{54124}, \country{Greece}}}

\abstract{
Electricity generated from renewable energy sources has been established as an efficient remedy for both energy shortages and the environmental pollution stemming from conventional energy production methods. Solar and wind power are two of the most dominant  renewable energy sources. The accurate forecasting of the energy generation of those sources facilitates their integration into electric grids, by minimizing the negative impact of uncertainty regarding their management and operation. This paper proposes a novel methodology for deterministic wind and solar energy generation forecasting for multiple  generation sites, utilizing multi-location weather forecasts. The method employs a U-shaped Temporal Convolutional Auto-Encoder (UTCAE) architecture for temporal processing of weather-related and energy-related time-series across each site. The Multi-sized Kernels convolutional Spatio-Temporal Attention (MKST-Attention), inspired by the multi-head scaled-dot product attention mechanism, is also proposed aiming to efficiently transfer temporal patterns from weather data to energy data, without a priori knowledge of the locations of the power stations and the locations of provided weather data. The conducted experimental evaluation on a day-ahead solar and wind energy forecasting scenario on five datasets demonstrated that the proposed method achieves top results, outperforming all competitive time-series forecasting state-of-the-art methods.}


\maketitle

\pagestyle{fancy} 
\fancyhf{} 
\fancyfoot[C]{This preprint has not undergone peer review or any post-submission improvements or corrections. The Version of Record of this article is published in Neural Computing and Applications, and is available online at https://doi.org/10.1007/s00521-024-10607-2} 

\section{Introduction}\label{introduction}
Fossil fuels, including coal, oil and natural gas, have long been the world's dominant sources of energy. Their fast depletion~\cite{hook2013fossil} as well as their overall impact in global warming~\cite{yoro20203emissions}, due to the emission of greenhouse gases, has led to the rapid increase in the exploitation of renewable energy sources, such as solar and wind energy, hydropower, geothermal energy and bioenergy. In recent years, solar radiation and wind have emerged as two major Renewable Energy Sources (RES), each accounting for a quarter of the total RES energy output~\cite{rahman2022res}. Two major drawbacks, encountered on both RES types, are that the corresponding power generation is variable and non-dispatchable~\cite{widen2015drawbacks}, i.e., the power generation output cannot (or have limited ability to) be adjusted to match the electricity demand. The integration of those RES types into existing power systems is mainly determined by: (i) the temporal variability of the output power from renewable energy generation stations and (ii) the accuracy of forecasts for the variable generation. Fluctuations in total non-dispatchable power generation along with fluctuations in the total power demand, require the in-parallel use of conventional dispatchable power generation methods on different time scales. Accurate forecasting of RES power generation~\cite{tzelepi2023deep} and the electric load demand~\cite{stentoumi2023anio}, can effectively mitigate the related uncertainties, contributing positively to the planning, management and operation of energy systems and thus achieving a lower level of use for conventional power generation methods.

Accurate renewable energy generation forecasting is a challenging task due to the intermittent and random nature of renewable energy data. Methods addressing the RES forecasting problem, can be mostly categorized as physical, statistical, machine learning and hybrid. Physical forecasting methods~\cite{lorenz2011regional}\cite{jorgensen2005honeymoon} focus on energy forecasting based on Numerical Weather Predictions (NWPs) and physical principles regarding the installed photovoltaic (PV) cells or wind turbines as well as the corresponding geographic formation. Physical methods do not require historical (past) information regarding   energy generation. Statistical forecasting methods, such as Autoregressive (AR) models ~\cite{rajagopalan2009wind}\cite{atique2019forecasting} and Markov chain models~\cite{carpinone2010very}\cite{bai2021improved}, aim to establish a mapping relationship between past-time and future-time energy generation time-series. Machine learning forecasting models, also referred as intelligent forecasting models in the literature, such as neural networks~\cite{dong2022_stcn_wind}\cite{hossain_2020_solar} and Support Vector Machines (SVMs)~\cite{li2020short}\cite{fan2018comparison}, attempt to establish a mathematical relationship between input and output through curve fitting and parameter optimization techniques~\cite{ahmed2019review}. Hybrid methods~\cite{aghajani2016novel}\cite{vrettos2019hybrid} incorporate a combination of different types of methods.

Based on the manifestation of the forecasting results, RES forecasting methods can also be categorized as deterministic~\cite{niu2020_wind}~\cite{rai_2022_solar} or probabilistic~\cite{7459241}~\cite{dowell2016aemo}. Deterministic methods output single-valued energy generation predictions for each time-step of the forecasting window. Probabilistic methods provide a wider view of possible energy generation outputs expressed as quantiles, prediction intervals (PIs), and distributions. 

Finally, RES forecasting methods can also be categorized based on the time horizon of the forecasting window. According to the relevant literature, the methods can be split into four categories~\cite{Hanifi2020}; ultra-short-term, short-term, medium-term, and long-term forecasting. However, these categories often remain ambiguous since no universally agreed classification criterion exists. Thus, the terms intra-hour, intra-day, day-ahead and week-ahead forecasting / prediction are more commonly encountered~\cite{hong2020review} in order to describe the forecast horizon in a more precise way. 

\begin{figure}[ht]
    \centering
    \includegraphics[width=\linewidth]{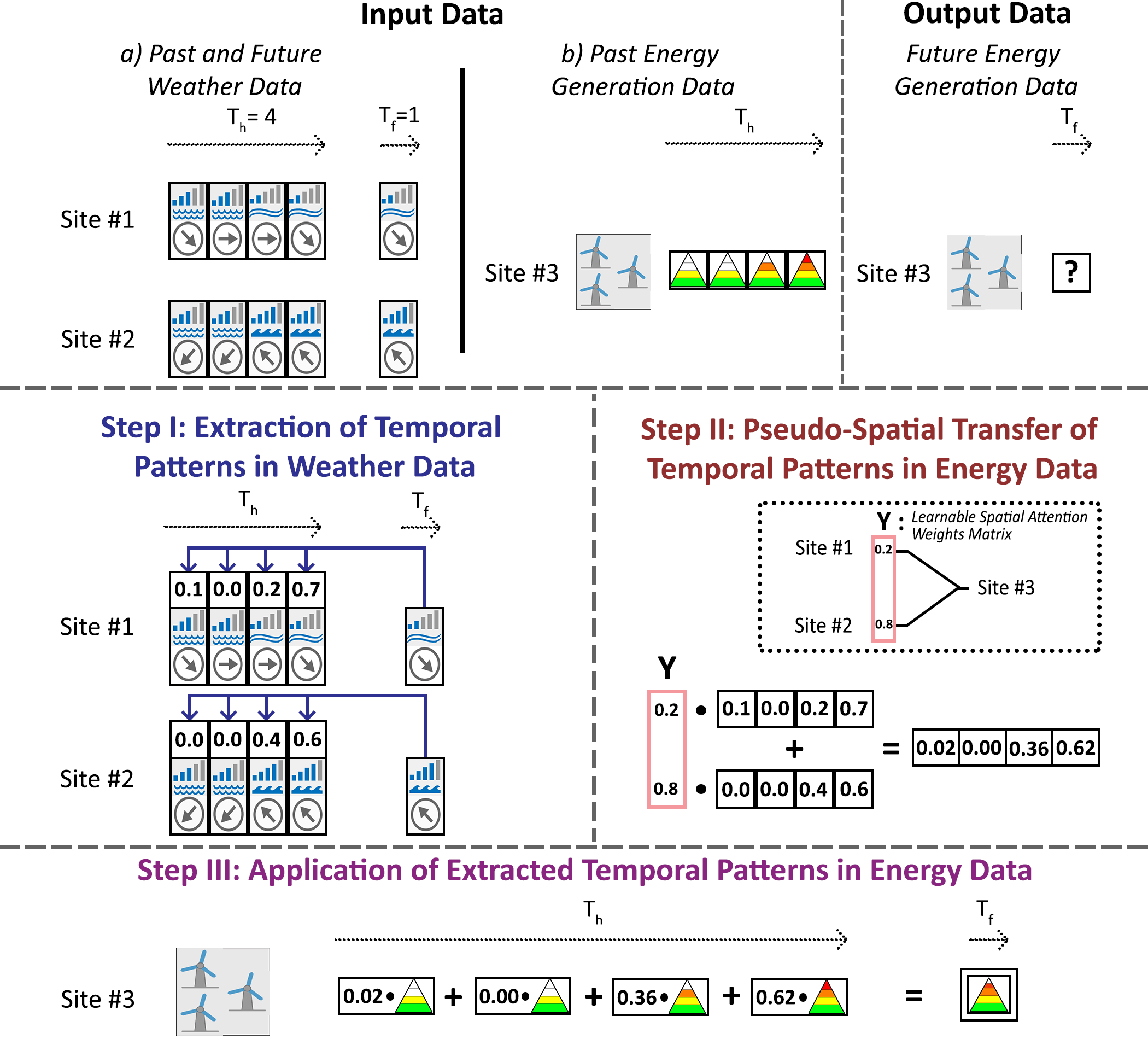}
    \caption{Abstract example of the Multi-sized Kernels convolutional Spatio-Temporal Attention (MKST-Attention) mechanism in the inference stage of a wind energy forecasting scenario, for a single wind power station (site \#3) and two  weather data locations (sites \#1 and \#2). The size of the future-time temporal window $T_f$ is 1 and the  size of past-time temporal window $T_h$ is 4.}  
    \label{fig:example}
\end{figure}

Our work, focuses mainly on deterministic machine learning-based wind and solar energy forecasting  approaches at power station level, that utilize weather data along with past-time energy measurements, as input. Specifically, we introduce a deterministic forecasting method for a such scenario, capable to simultaneously generate multi-step (i.e., for multiple time instances) energy production forecasts for a given set of power stations. In particular, we employ a U-shaped Temporal Convolutional Auto-Encoder (UTCAE) for temporal processing of weather and energy data across each site. The motivation behind this choice 
is to enrich the representations of a given time-series, by exploring patterns across various temporal resolutions, thus capturing both global and local contextual information. In addition, we propose the use of a Multi-sized Kernels convolutional Spatio-Temporal Attention (MKST-Attention) mechanism that is responsible for transferring temporal patterns from weather data to energy data, without a priori knowledge of the locations of the power stations and the locations where weather data were measured (or forecasted). For each energy generation site, the mechanism is able to discover the weather data locations which are significant for the  site (i.e., they affect the site) and transfer temporal patterns from the corresponding weather data subset to the energy generation data. Figure~\ref{fig:example}  illustrates an instance of MKST-Attention during the inference stage. Overall, the main contributions of this work can be summarized as follows:
\begin{itemize}
  \item We propose a deterministic multi-step wind and solar energy forecasting methodology for multiple energy generation sites, capable to efficiently utilize weather data from multiple sites, without a priori knowledge of the location of either the energy generation or the weather data sites.
  \item Inspired by the Multi-head Scaled-dot Product Attention mechanism, we propose the Multi-sized Kernels convolutional Spatio-Temporal Attention mechanism, which can explore spatio-temporal patterns between two sets of time-series data.
  \item We evaluate the proposed methodology on three wind energy forecasting datasets and two solar energy forecasting datasets, showcasing its effectiveness. 
\end{itemize}

The proposed method is an extension of the method in \cite{symeonidis2023wind}, which was designed specifically for deterministic wind energy forecasting. Both methods utilize a similar convolutional attention mechanism for exploring spatio-temporal relations between energy generation data and weather data. A major difference of the proposed method with the method in \cite{symeonidis2023wind}, lies on the computation of the spatial relations between the energy generation sites and the sites of weather data. In \cite{symeonidis2023wind}, the spatial relations are time-varying and are computed based on the respective energy and weather data fed as input, whereas in the method proposed in this paper  the  spatial relations  explored by the attention mechanism (named Multi-sized Kernels convolutional Spatio-Temporal Attention) remain fixed after the corresponding training process. Apart from the differences associated to the implemented attention mechanism, the method in \cite{symeonidis2023wind} doesn't incorporate a mechanism to temporally process weather data, adopting a rather shallow architecture. In contrast, the current approach employs the U-shaped Temporal Convolutional Auto-Encoder (UTCAE) and incorporates a scheme of repeated Joint Processing Blocks (JPBs) with the objective of generating contextually informative representations for both the energy-related and weather-related time-series data.

\section{Previous Work}\label{Sec:rel_work}

In this Section, we discuss previous work on deterministic wind and solar energy forecasting at power station level, focusing on deep learning methods. A brief summary of the discussed methods is provided in Table~\ref{Tab:related_methods}.

\subsection{Deterministic Wind Energy Forecasting Methods for Single/Multiple Power Station(s)}

In \cite{zhu2020_tcn_wind}, the authors suggested the use of a Temporal Convolutional Neural Network (TCNN), aiming to achieve accurate single station short-term wind energy forecasts. The method's performance was evaluated on a simulated dataset, provided by the National Renewable Energy Laboratory (NREL), achieving top results against the rest of the baseline methods. In addition, the evaluation highlighted the stable learning process and the strong generalization ability of the proposed method. In \cite{dong2022_stcn_wind}, a Spatio-Temporal Convolutional Neural Network (STCNN) was employed in a multiple-stations short-term wind energy forecasting scenario. According to the authors, the method's performance gains were due to  its architecture, which combined the structure of directed Graph Convolutional Network (GCNs) with that of TCNNs, allowing the effective capture of asymmetric spatial correlations at different temporal scales. Similarly, the authors in \cite{qu_2022_tr_wind} applied the Transformer model in a short-term multiple-stations wind energy forecasting scenario. The employed model was able to extract different levels of correlation between wind farms and provide accurate wind power forecasting results. In \cite{liu2019_lstm_wind}, the authors proposed a short-term wind energy forecasting method based on Discrete Wavelet Transform (DWT) and Long Short-Term Memory (LSTM) networks. The method adopts a divide and conquer
strategy, in which DWT is used to decompose original wind energy data into sub-signals, while several independent LSTMs are employed to approximate the temporal dynamic behaviors of these  sub-signals. The proposed method achieved top prediction accuracy rates against other state-of-the-art methods in a single-farm short-term wind energy forecasting scenario.  In \cite{niu2020_wind}, the authors designed a sequence-to-sequence model for a multi-step-ahead single-station wind energy forecasting scenario. The model architecture consists of two groups of Attention-based Gated Recurrent Unit (AGRU) blocks, employed in an encoder-decoder architecture.  In the proposed model, the correlation between different forecasting tasks was considered in the GRU and an attention mechanism helped select important features. An experimental evaluation, regarding the accuracy, the computational efficiency and the feature selection capabilities of the proposed model demonstrated its superiority against other state-of-the-art methods.

\subsection{Deterministic Solar Energy Forecasting Methods for Single/Multiple Power Station(s)}

In \cite{lin_2020_tcn_solar}, the authors studied the use of a TCNN in a day-ahead forecasting scenario with half-hour temporal resolution. According to the authors, the model was able to maintain a much longer effective history, compared to  competing methods, by extracting informative features from long sequences using dilated causal convolutional filters. The authors in~\cite{jung_2020_solar} proposed a monthly PV power generation forecasting method based on the LSTM model, using meteorological and energy data collected from existing power stations, aiming to predict the energy generation of a potential power station at a new location.
Through experiments using historical training data from 134 solar power stations, the proposed approach was able to perform well in a testing set containing time-series of 30 solar power stations not used in the training process. In \cite{rai_2022_solar}, the authors developed a single-station solar energy forecasting method based on the fusion of a sequence-to-sequence auto-encoder and a gated recurrent unit. While the GRU makes use of the data’s time dependencies, the auto-encoder extracts the correlation of significant parameters like solar irradiance. The proposed method was able to perform better than
LSTM and CNN-LSTM-based models utilizing a variety of performance metrics, including Mean Square Error (MSE) and Mean
Absolute Error (MAE). The authors in \cite{jeong2019multi}, proposed a spatio-temporal CNN for solar energy forecasting, which exploits the spatial information of multiple power stations. First, the proposed Greedy Adjoining Algorithm (GAA) constructs a space-time matrix that rearranges the solar energy generation sites based on their geographical proximity. Then, a CNN is applied to learn spatio-temporal relations. The conducted experimental evaluation showed that the proposed STCNN achieves the highest error reduction when data from multiple power stations are aggregated. In \cite{simeunovic2022interpretable}, the authors introduced a novel spatio-temporal Graph Attention network (GAT) for solar energy forecasting in multiple power stations. The proposed method can capture different dynamical spatio-temporal correlations for different parts of the forecasting horizon. Thus, it is possible to interpret which power stations have the most influence when performing short-, medium- and long-term intra-day forecasts. The proposed method outperformed state-of-the-art methods in 4 to 6 hours ahead solar energy forecasting scenarios in all employed evaluation metric, on real and synthetic datasets.

\begin{center}
\begin{table}[!ht]
\begin{tabular}{|c|c|c|c|}
\hline
\multirow{3}{*}{\textbf{Method}} &
\multirow{2}{*}{\textbf{Energy}} &
\multirow{3}{*}{\textbf{Architecture}} &
\textbf{Single/Multiple} \\ 
& \multirow{2}{*}{\textbf{Type}} & & \textbf{Station(s)} \\
& & & \textbf{Forecasting}  \\ \hline

 \multicolumn{1}{|c|}{\cite{liu2019_lstm_wind}} & \multicolumn{1}{c|}{wind} & \multicolumn{1}{c|}{RNN, DWT} & single \\ \hline
 
\multicolumn{1}{|c|}{\cite{zhu2020_tcn_wind}} & \multicolumn{1}{c|}{wind} & \multicolumn{1}{c|}{TCNN} & single \\ \hline

\multicolumn{1}{|c|}{\cite{niu2020_wind}} & \multicolumn{1}{c|}{wind} & \multicolumn{1}{c|}{AGRU} & single \\ \hline

\multicolumn{1}{|c|}{\cite{dong2022_stcn_wind}}& \multicolumn{1}{c|}{wind} & \multicolumn{1}{c|}{STCNN, GCN} & multiple \\ \hline

\multicolumn{1}{|c|}{\cite{qu_2022_tr_wind}} & \multicolumn{1}{c|}{wind} & \multicolumn{1}{c|}{Transformer} & multiple \\ \hline

\multicolumn{1}{|c|}{\cite{symeonidis2023wind}} & \multicolumn{1}{c|}{wind} & \multicolumn{1}{c|}{STAN} & multiple \\ \hline

\multicolumn{1}{|c|}{\cite{lin_2020_tcn_solar}} & \multicolumn{1}{c|}{solar} & \multicolumn{1}{c|}{TCNN} & single \\ \hline

\multicolumn{1}{|c|}{\cite{rai_2022_solar}} & \multicolumn{1}{c|}{solar} & \multicolumn{1}{c|}{Auto-Encoder,GRU} & single  \\ \hline

\multicolumn{1}{|c|}{\cite{jung_2020_solar}} & \multicolumn{1}{c|}{solar} & \multicolumn{1}{c|}{LSTM} & multiple  \\ \hline

\multicolumn{1}{|c|}{\cite{jeong2019multi}} & \multicolumn{1}{c|}{solar} & \multicolumn{1}{c|}{STCNN} & multiple  \\ \hline
\multicolumn{1}{|c|}{\cite{simeunovic2022interpretable}} & \multicolumn{1}{c|}{solar} & \multicolumn{1}{c|}{GAT} & multiple  \\ \hline

\end{tabular}
\caption{\centering Brief summary of the discussed deterministic wind and solar energy forecasting methods.}
\label{Tab:related_methods}
\end{table}
\vspace{-0.4cm}
\end{center}

\section{Proposed Method}\label{prop_method}
Our wind/solar energy forecasting approach is inspired by attention-based and temporal convolutional architectures, both commonly encountered in the  task of generic time-series forecasting. The core mechanism of the proposed method, named Multi-sized Kernels convolutional Spatio-Temporal Attention (MKST-Attention), is able to spatially transfer temporal patterns from  weather data time-series to energy generation time-series. In addition, aiming to temporally process both weather and energy generation data across sites, we employ a U-shaped Temporal Convolutional Auto-Encoder (UTCAE), capable to capture multi-scale contextual information along the temporal dimension. To present the proposed method in an organized and clear way, the rest of the section is structured as follows. Section~\ref{subsec:prop_statement} states the corresponding problem and explains the adopted notations. Subsequently, Section~\ref{sec:mkcsdpa} presents the Multi-sized Kernels convolutional Attention (MK-Attention), a novel variant of multi-head attention. In Sections~\ref{sec:mksta} and~\ref{sec:uced} the MKST-Attention mechanism and the architecture of the UTCAE are presented, respectively. Both are employed by the Joint Processing Block (JPB), presented in Section~\ref{sec:jpb}, aiming to process and refine the latent representations of energy generation  and weather data. Finally, the overall architecture of the proposed method is presented in detail in Section~\ref{sec:architecture}.

\subsection{Problem Statement and Notations}
\label{subsec:prop_statement}
The problem of wind and solar energy forecasting, addressed in this paper, can be formulated as:
\begin{equation}
\hat{\textbf{E}}^{f} = g(\textbf{E}^{h}, \textbf{W}^{h}, \textbf{W}^{f}) 
\end{equation}
In this equation $\textbf{E}^{h}$ $\in \mathbb{R}^{L_E \times T_h \times 1}$ corresponds to the past/history ($h$: history) energy generation measurements from $L_E$ locations (e.g., locations of wind farms, PV power stations, etc.,), where $T_h$ is the size of the past-time temporal window. Moreover, $\textbf{W}^{h}$ $\in \mathbb{R}^{L_W \times T_h \times D_W}$ corresponds to past-time weather data from $L_W$ distinct locations, where $D_W$ is the number of  of weather data measurements/variables.  Also, $\textbf{W}^{f}$ $\in \mathbb{R}^{L_W \times T_f \times D_W}$ corresponds to future-time ($f$:future) weather data, i.e. weather forecasts, where $T_f$ is the size of the future-time temporal window. Finally, $\hat{\textbf{E}}^{f}$ $\in \mathbb{R}^{L_E \times T_f \times 1}$ corresponds to the wind or solar energy predictions that are generated by the method simultaneously for each of the $L_E$ locations.


In order to ensure clarity and consistency in representing matrix or tensor indexing throughout this work, we adopt the following notation scheme:  matrix or tensor elements will be denoted using indices enclosed within angle brackets ($\langle .  \rangle$). For instance, elements of a third-order tensor $\textbf{X} \in \mathbb{R}^{N_1 \times N_2 \times N_3}$, will be referenced as $\textbf{x} \langle i, j, k \rangle $ where $i$, $j$, and $k$ represent indices along the first, second, and third dimension, respectively. In addition, the  dot symbol ($\cdot$)  will represent the selection of all indices within the specified dimension. For instance, the notation $\textbf{X} \langle i,\cdot, \cdot \rangle $ denotes the selection of  $\textbf{X}$ elements  with index $i$ along the first dimension and all index values across the second and third dimensions. These notations will be consistently employed in all subsequent discussions and equations involving matrices or tensors.

In addition, let  $\textbf{B} \in \mathbb{R}^{N_1 \times N_2 \times N_3}$ and $\textbf{D} \in \mathbb{R}^{N_1 \times N_3 \times N_4}$ be two batches of $N_1$ matrices, where $\textbf{B} = [\textbf{B}_1, ... , \textbf{B}_{N_1}]$, and   $\textbf{D} = [\textbf{D}_1, ... , \textbf{D}_{N_1}]$.  We can then define Batch Matrix-Multiplication (BMM) as $\textbf{B} \otimes \textbf{D} = \textbf{G}$, where  $\textbf{G} \in \mathbb{R}^{N_1 \times N_2 \times N_4}$ is  a batch of $N_1$ matrices,  $\textbf{G} =[\textbf{G}_1, ... , \textbf{G}_{N_1}]$.

Adopting the typical attention mechanism \cite{shih2019temporal}, the energy generation prediction $\hat{e}^f \langle s, t \rangle $ at location $s$ and time-step $t$, namely one of the elements of $\hat{\textbf{E}}^f= [[\hat{e}^f\langle 1, 1 \rangle, ..., \hat{e}^f\langle 1, T_f \rangle], ...,  [\hat{e}^f\langle L_E, 1 \rangle, ..., \hat{e}^f\langle L_E, T_f \rangle]]$ can be formulated as:

\vspace{-0.2cm}
\begin{equation}
\label{eq:idea_mt}
\hat{e}^{f} \langle s, t \rangle = (\sum_{j=1}^{T_h} \alpha\langle s,t,j \rangle \textbf{E}^{h,r} \langle s, j, \cdot \rangle ) \textbf{C}  
\end{equation}

\begin{equation}\notag
\text{where  } \sum_{j=1}^{T_h} \alpha \langle s, t,j \rangle = 1 
\end{equation}

\begin{equation}\notag
\forall s \in [1,..,L_E], \forall t \in [1,..,T_f],  \forall j \in [1, ... T_h]
\end{equation}

\noindent In the above formulas, $\textbf{A} \in \mathbb{R}^{L_E \times T_f \times T_h}$ corresponds to a non-negative attention weights tensor that expresses future-time energy generation latent representations of each site, as linear combinations of past-time energy generation latent representations of the same site. 
In addition, $\textbf{E}^{h,r} \langle s, j, \cdot \rangle \in \mathbb{R}^{D_{r}}$ corresponds to the latent representation ($r$: representations) of the past energy generation measurement at location $s$ and past time-step $j$ and $D_r$ denotes the size of each energy generation latent representation. Moreover, $\textbf{C} \in  \mathbb{R}^{D_r\times 1}$ are learnable parameters of a linear operator. In this formulation, wind or solar energy generation at $s$-th location and $t$-th time-step is predicted based on the temporal patterns, imposed by attention weights $\textbf{A} \langle s, t , \cdot \rangle$, and past energy generation measurements (more specifically their latent representations) within the respective temporal window. Our objective is to explore, the previously described, pseudo-spatial and temporal relations between $\textbf{E}^h$, $\textbf{W}^h$ and $\textbf{W}^f$ in order to efficiently approximate \textbf{A}.

\begin{figure}[ht]
\centering
\begin{subfigure}{.40\textwidth}
  \centering
  \includegraphics[width=\linewidth]{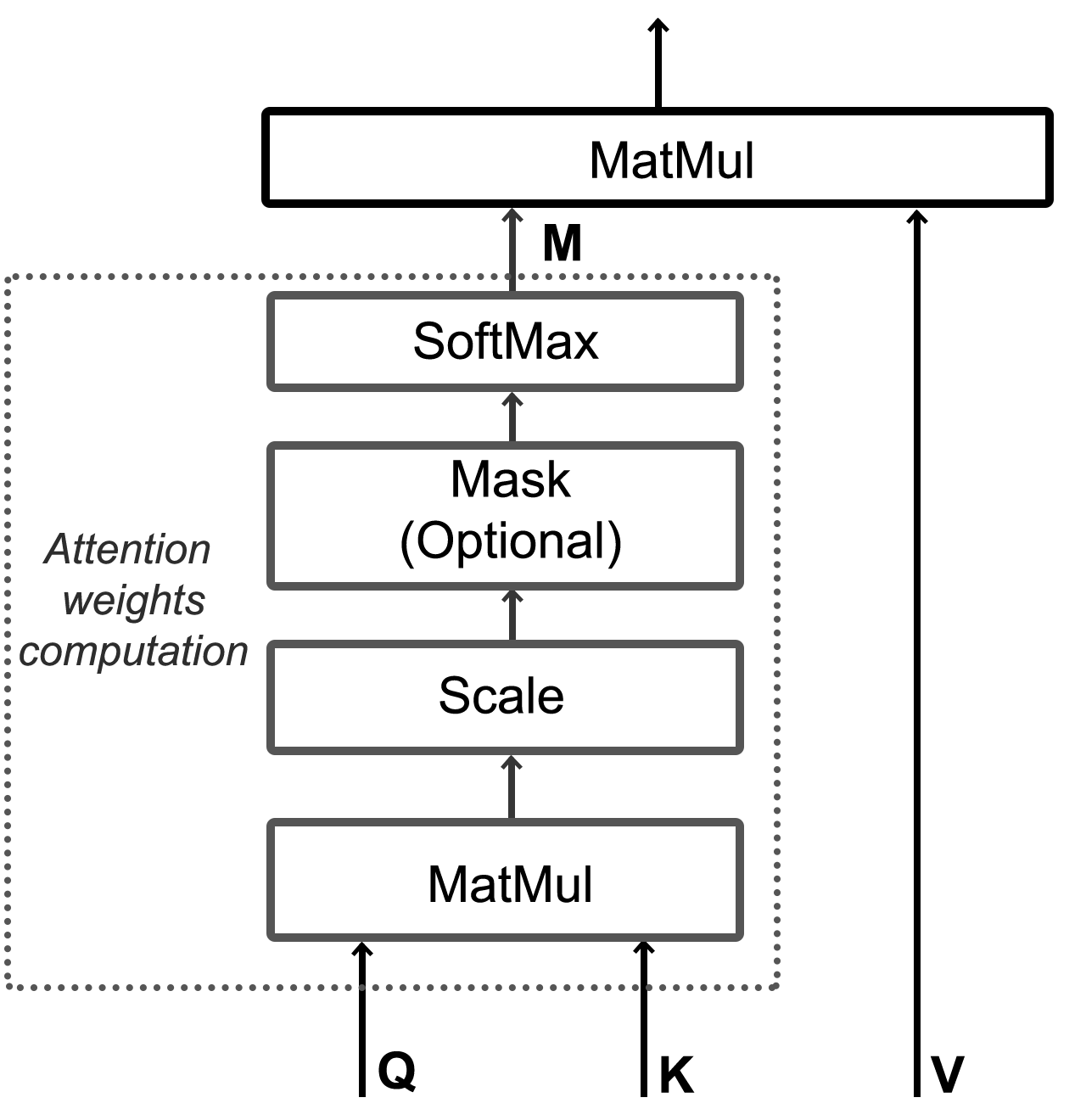}
  \caption{\centering Scaled Dot-Product Attention}
  \label{fig:sdpa}
\end{subfigure}
\begin{subfigure}{.55\textwidth}
  \centering
  \includegraphics[width=\linewidth]{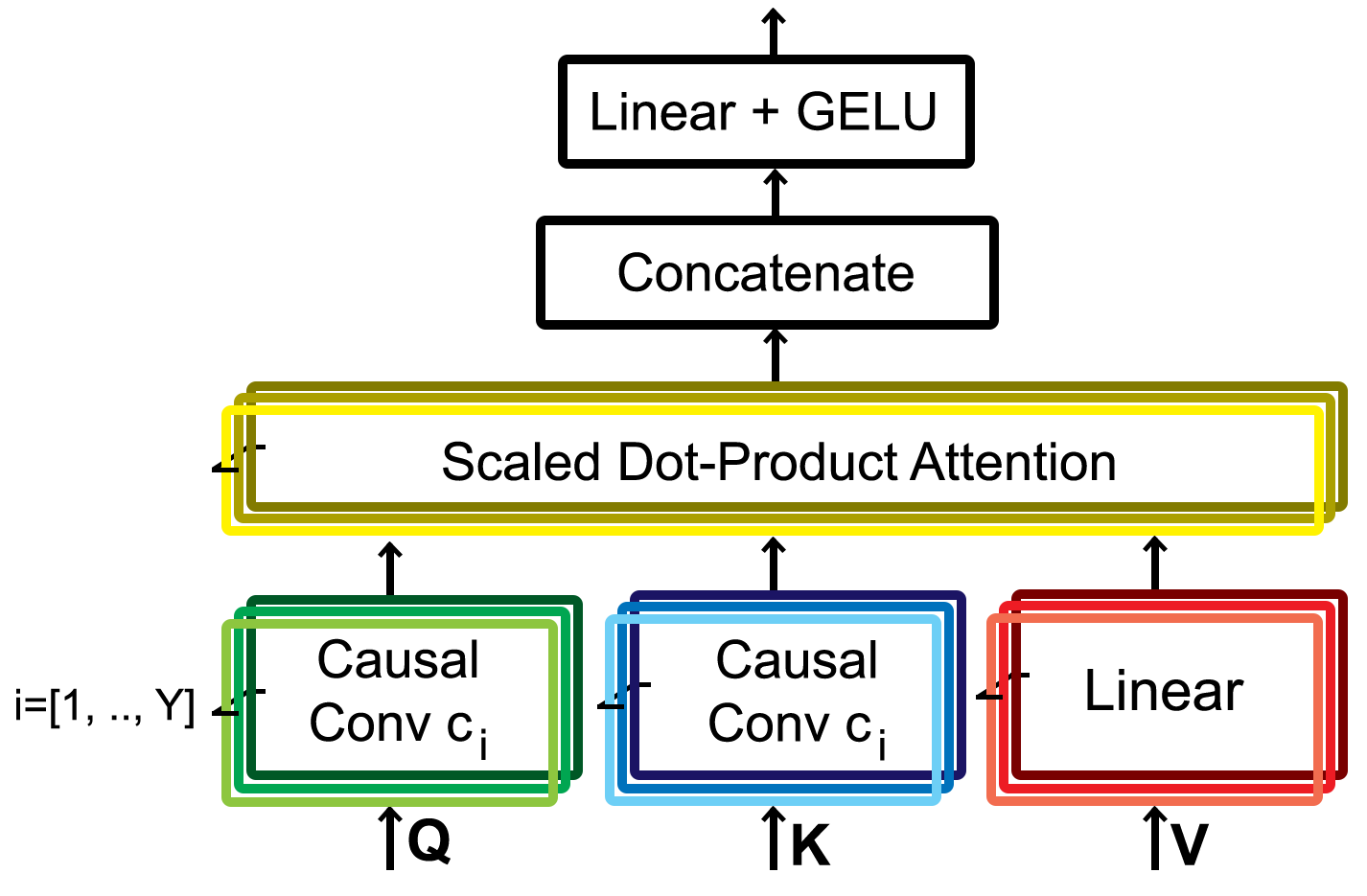}
  \caption{\centering Multi-sized kernels convolutional scaled dot-product attention.}
  \label{fig:mkcsdpa}
\end{subfigure}
\caption{(a):  Scaled Dot-Product Attention, (b): the novel  Multi-sized kernels convolutional scaled dot-product attention. $c_i$ denotes the size of the $i$-th  convolutional kernel, employed in the temporal domain whereas $\Upsilon$ denotes the number of convolutional kernels.}
\label{fig:attention}
\end{figure}

\subsection{Multi-Sized Kernels Convolutional Scaled Dot-Product Attention}
\label{sec:mkcsdpa}

The Scaled Dot-Product Attention was presented in \cite{transformer2017} and is formulated as follows:

\begin{equation}
\label{eqn:attention}
    Attention(\textbf{Q},\textbf{K},\textbf{V}) = \textbf{M} \textbf{V} \text{, \ \ }
\end{equation}
\begin{equation}
\label{eqn:atten_scores}
    \text{where \ } \textbf{M} =     softmax(\frac{\textbf{QK}^{T}}{\sqrt{D_K}})
\end{equation}

\noindent $\textbf{Q} \in \mathbb{R}^{N_Q \times D_K}$,  $\textbf{K} \in \mathbb{R}^{N_K \times D_K}$  and $\textbf{V} \in \mathbb{R}^{N_K \times D_V}$ are the queries, keys and values respectively. Queries and  keys have a dimension of $D_K$, while  values have a dimension of $D_V$. $N_Q$ is the number of queries while $N_K$ is the number of keys and values. An illustration of the mechanism is depicted in Figure~\ref{fig:sdpa}. Multi-head attention was also proposed in \cite{transformer2017}, allowing various attention mechanisms, including scaled dot-product attention, to run in parallel. To this end, instead of performing a single attention computation on queries, keys, and values with a universal dimension of $D_r$, the authors proposed their transformation with $\Upsilon$ independently learned linear projections. The attention computation is then performed, in parallel, on those $\Upsilon$ projected queries, keys, and values. More specifically, the multi-head attention module can be formulated as:

\vspace{-0.2cm}
\begin{equation}
\label{eqn:multihead}
MultiHead(\textbf{Q},\textbf{K},\textbf{V}) = [\textbf{p}_1,...,\textbf{p}_{\Upsilon}]\textbf{S}^O,
\end{equation}

\begin{equation}
\label{eqn:multihead_t}
\text{where \ } \textbf{p}_i = Attention(\textbf{Q}\textbf{S}^{Q}_i, \textbf{K}\textbf{S}^{K}_i, \textbf{V}\textbf{S}^{V}_i)
\end{equation}

\noindent In this formulation, $\textbf{S}^{Q}_i \in \mathbb{R}^{D_r \times D_K}$, $\textbf{S}^{K}_i \in \mathbb{R}^{D_r \times D_K}$, $\textbf{S}^{V}_i \in \mathbb{R}^{D_r \times D_V}$, $\textbf{S}^{O} \in \mathbb{R}^{\Upsilon D_V \times D_r}$ are projection parameter matrices, $\Upsilon$ is the number of heads, $D_K=D_V=\frac{D_r}{\Upsilon}$, and the operator $[...]$ implies concatenation.

On the original formulation, the scaled dot-product attention was designed to explore point-wise similarities between queries and keys. However, in most time-series analysis tasks, information regarding the  context surrounding the observed points is vital for exploring patterns among the series. The authors in \cite{NEURIPS2019_6775a063} were able to employ convolutions of kernel size $c$ to transform inputs into queries and keys. Thus, local context was exploited in the query-key matching, improving the way temporal patterns among the corresponding time-series are explored. The authors experimented with various values of $c$ in order to find the optimal one. Instead of selecting one kernel size, we propose the Multi-sized Kernels convolutional scaled dot-product Attention, which can be formulated by modifying Equations~\ref{eqn:multihead} and \ref{eqn:multihead_t} as follows:

\vspace{-0.2cm}
\begin{equation}
\label{eqn:multikernel}
MultiKernel(\textbf{Q},\textbf{K},\textbf{V}) = [\textbf{p}_1,...,\textbf{p}_\Upsilon]\textbf{S}^O,
\end{equation}

\begin{equation}
\label{eqn:att_kernels}
\textbf{p}_i = Attention(g_Q(\textbf{Q}, c_i), g_K(\textbf{K}, c_i), \textbf{V}\textbf{S}^{V}_i)
\end{equation}

\noindent where $g_Q(\textbf{Q}, c)$ and $g_K(\textbf{K}, c)$ correspond to  1-D convolutional layers of $D_V$ kernels, $c$ being the size of 1-D convolutional kernels. In this formulation, convolutions with $\Upsilon$ different kernel sizes are applied on \textbf{Q}, \textbf{K} resulting in $\Upsilon$ heads. In addition, a separate linear projection is applied on \textbf{V} for each head.
In this way, the scaled dot-product attention is computed separately for each head. It should be noted that this process cannot be executed simultaneously but iterativelly for all heads. Then, the outputs are concatenated and projected as depicted in Figure~\ref{fig:mkcsdpa}. 

\subsection{Multi-Sized Kernels Convolutional Spatio-Temporal Attention}
\label{sec:mksta}

Aiming to formulate the spatio-temporal attention (ST-Attention) we model queries, keys and values in space-time as $\textbf{Q} \in \mathbb{R}^{L_K \times T_Q \times D_K}$, $\textbf{K} \in \mathbb{R}^{L_K \times T_K \times D_K}$, $\textbf{V} \in \mathbb{R}^{L_V \times T_K \times D_K}$, respectively, where  the first dimension of these tensors is the spatial one, the second dimension is the temporal one  and the last corresponds to the size of each vector representation. In addition, we define $\textbf{Y} \in \mathbb{R}^{ L_V \times L_K}$, which is a matrix that holds the spatial relations between the $L_V$ sites of the values and the $L_K$ sites of queries and keys. Based on Equation~\ref{eqn:attention}, we formulate spatio-temporal attention (ST-Attention) as follows:

\begin{equation}
\label{eqn:attention_b}
    ST\text{-}Attention(\textbf{Q}, \textbf{K},\textbf{V},  {\textbf{Y}) =  \textbf{B} \otimes \textbf{V} \ \ }
\end{equation}

\noindent where
\begin{equation}
\label{eqn:atten_scores_1}
     \textbf{B} = [ \sum_{i=1}^{L_K} y \langle 1,i \rangle \textbf{M} \langle i, \cdot, \cdot \rangle, ... , \sum_{i=1}^{L_K} y \langle L_V, i \rangle  \textbf{M} \langle i, \cdot, \cdot \rangle] \text{, \ \ }
\end{equation}

\begin{equation}
\label{eqn:atten_scores_2}
     \textbf{M}\langle i, \cdot, \cdot \rangle = softmax(\frac{\textbf{Q}\langle i, \cdot, \cdot \rangle \textbf{K}\langle i, \cdot, \cdot \rangle ^{T}}{\sqrt{D_K}}) \text{, }  
\end{equation}

\begin{equation}
    \sum_{i=1}^{L_K} y \langle j, i \rangle = 1 \text{, \ \ } 
\end{equation}

\begin{equation}\notag
\forall i \in [1, L_K], \forall j \in [1, L_V] 
\end{equation}

\noindent In this formulation $\textbf{M} \in \mathbb{R}^{L_K \times T_Q \times T_K}$ holds temporal relations for each of the $L_K$ sites of keys/values separately. More specifically, $\textbf{M} \langle i, \cdot, \cdot \rangle$ holds temporal relations formed between $\textbf{Q} \langle i, \cdot, \cdot \rangle$ and $\textbf{K} \langle i, \cdot, \cdot \rangle$ for the $i$-th site of queries and keys. The temporal relations in $\textbf{M}$ can be spatially applied to the sites of the values through $\textbf{Y}$, forming tensor $\textbf{B}  \in \mathbb{R}^{L_V \times T_Q \times T_K}$, as shown in Equation~\ref{eqn:atten_scores_1}. Finally, the output of ST-Attention is computed as the batch matrix-multiplication of tensors $\textbf{B}$ and $\textbf{V}$.

Combining the MK-Attention, expressed in Equations~\ref{eqn:multikernel},~\ref{eqn:att_kernels}  with ST-Attention, we formulate the Multi-sized Kernels convolutional Spatio-Temporal Attention (MKST-Attention):

\begin{equation}
\label{eqn:mksta}
    MKST\text{-}Attention(\textbf{Q}, \textbf{K},\textbf{V}, \textbf{Y}) = [\textbf{p}_1,...,\textbf{p}_{\Upsilon}] \otimes \textbf{S}^A
\end{equation}

\begin{equation}
\label{eqn:mksta_kernels}
\textbf{p}_i = ST\text{-}Attention(g_Q(\textbf{Q}, c_i), g_K(\textbf{K}, c_i), \textbf{V}\textbf{S}^{V}_i, \textbf{Y})
\end{equation}

\noindent where $\textbf{S}^A  \in \mathbb{R}^{L_V \times \Upsilon D_V \times D_r}$ is a tensor formed as a batch of $L_V$ repetitions of the $\textbf{S}^O \in \mathbb{R}^{\Upsilon D_V \times D_r}$ projection parameter matrix.

\subsection{U-Shaped Temporal Convolutional Auto-Encoder}
\label{sec:uced}

\begin{figure}[H]
    \centering
    \includegraphics[width=\linewidth]{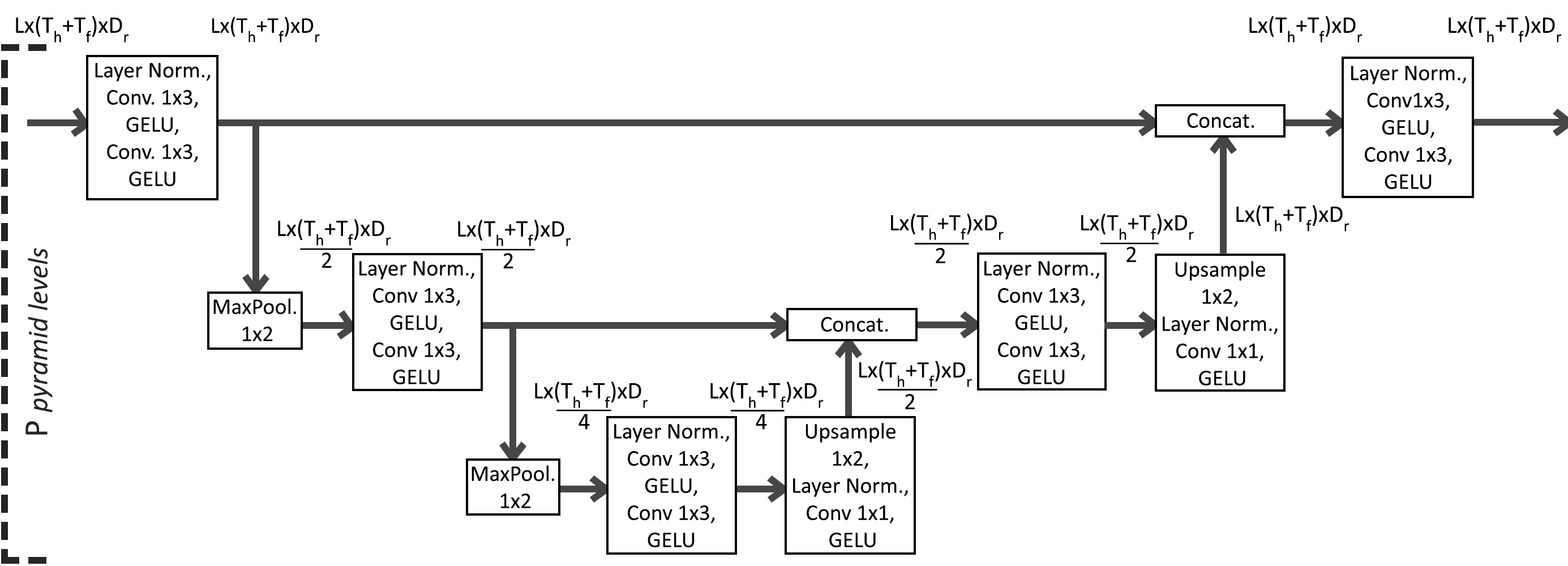}
    \caption{Illustration of the architecture of the U-shaped Temporal Convolutional Auto-Encoder, where the number of pyramid levels is 3.}
    \label{fig:unet}
\end{figure}

High temporal resolution  weather  and energy generation data (e.g. per minute or hour measurements/estimations) might contain fluctuations or noise resulting  from sensor inaccuracies or due to  their processing procedure. As a result, high-frequency changes in weather data might not coincide with equivalent changes in energy generation data and vice versa. This has a very negative impact on identifying temporal patterns between those two modalities. Aiming to diminish such data inaccuracies as well as to extract essential contextual information from multiple temporal scales 
we employ a U-shaped Temporal Convolutional Auto-Encoder (UTCAE). The corresponding architecture, depicted in Figure~\ref{fig:unet}, shares similarities with the architecture of U-Net~\cite{ronneberger2015u}, a neural network originally proposed for the task of biomedical image segmentation. It should be noted that UNet-inspired architectures~\cite{perslev2019u}~\cite{madhusudhanan2022u} have also been applied in time-series related tasks. The architecture of UTCAE, has a pyramid structure of $P$ levels, and consists of two parts.
The encoding part  comprises mostly of convolutional layers followed by max pooling operations, aiming to progressively reduce the temporal dimension of the time-series data while extracting latent representations with more contextual information. The decoding part, at its core, consists of upsampling operations, thus increasing  gradually the temporal dimension back to the original size. Skip connections, enable information from the layers of the encoder to be delivered directly to the decoder, facilitating the gradients to propagate smoother during training by providing shortcuts through the network.

\subsection{Joint Processing Block}
\label{sec:jpb}

The Joint Processing Block (JPB) can be formulated as: 

\begin{equation}
   \textbf{E}^{out}, \textbf{W}^{out} = \textit{JPB}(\textbf{E}^{in}, \textbf{W}^{in}, \textbf{Y})  
\end{equation}

where

\begin{equation}
    \textbf{E}^{out} =MKST\text{-}Attention(\textbf{W}^{out}, \textbf{W}^{out},\textbf{E}^{u}, \textbf{Y}) + \textbf{E}^{u} \text{,}
\end{equation}
\begin{equation}
     \textbf{W}^{out} = \textit{U}_W(\textbf{W}^{in}) \text{,}
\end{equation}
\begin{equation}
     \textbf{E}^{u} = \textit{U}_E(\textbf{E}^{in}).
\end{equation}

\begin{figure}[H]
    \centering
\includegraphics[width=0.75\linewidth]{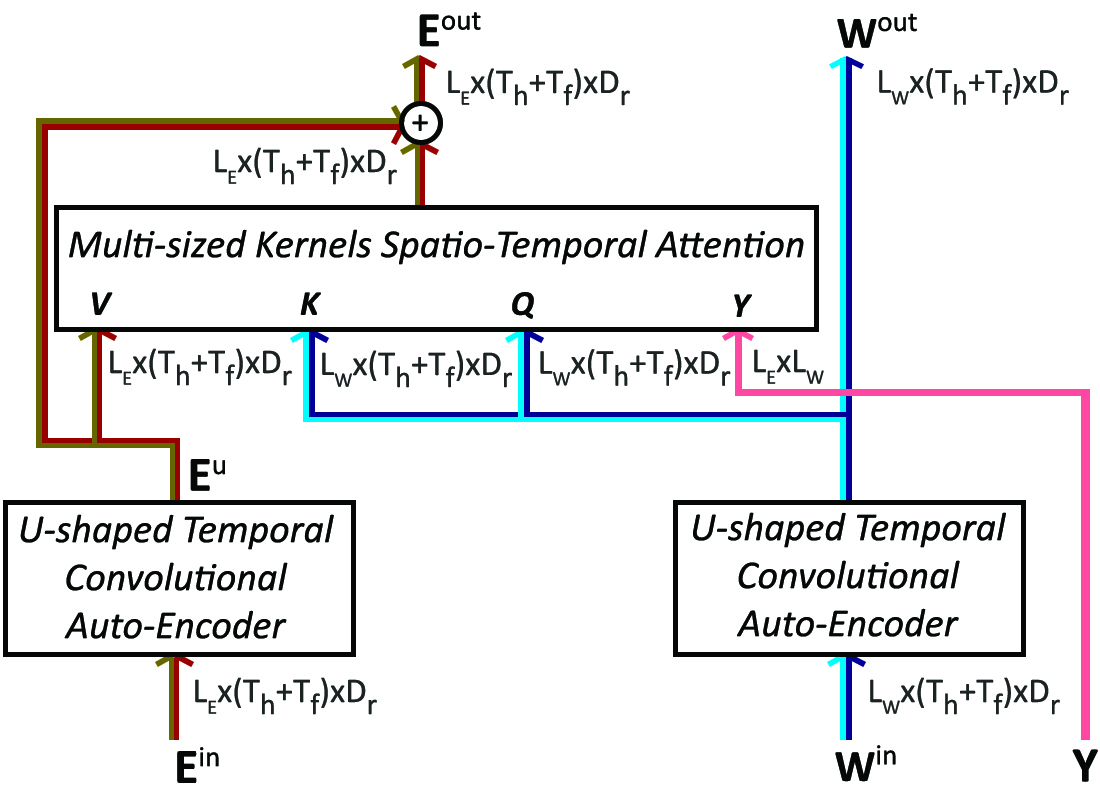}
    \caption{Architecture of the Joint Processing Block.}
    \label{fig:jpb}
\end{figure}

As input, the block receives spatio-temporal energy generation data representations $\textbf{E}^{in} \in \mathbb{R}^{L_E \times (T_h + T_f) \times D_r}$ and spatio-temporal weather data representations $\textbf{W}^{in} \in \mathbb{R}^{L_W \times (T_h + T_f) \times D_r}$. Matrix $\textbf{Y} \in \mathbb{R}^{L_E \times L_W}$, which holds the spatial relations between the $L_E$ energy generation sites and the $L_W$ weather data sites, is also provided to the JPB. At first stage, the energy generation data and weather data are separately processed by the U-shaped Temporal Convolutional Auto-Encoders, denoted as $U_E(*)$ and $U_W(*)$, resulting in refined energy- and weather-data latent representations $\textbf{E}^{u} \in \mathbb{R}^{L_E \times (T_h + T_f) \times D_r}$ and $\textbf{W}^{out} \in \mathbb{R}^{L_W \times (T_h + T_f) \times D_r}$. Then, those  latent representations, along with $\textbf{Y} \in \mathbb{R}^{L_E \times L_W}$ are fed to MKST-Attention, which outputs the final energy generation representation $\textbf{E}^{out} \in \mathbb{R}^{L_E \times (T_h + T_f) \times D_r}$. A residual connection is also applied between $\textbf{E}^{out}$ and $\textbf{E}^{u}$. The architecture of JPB is  depicted in Figure~\ref{fig:jpb}.

\subsection{Overall Architecture}
\label{sec:architecture}

\begin{figure}[H]
    \centering
    \includegraphics[width=\linewidth]{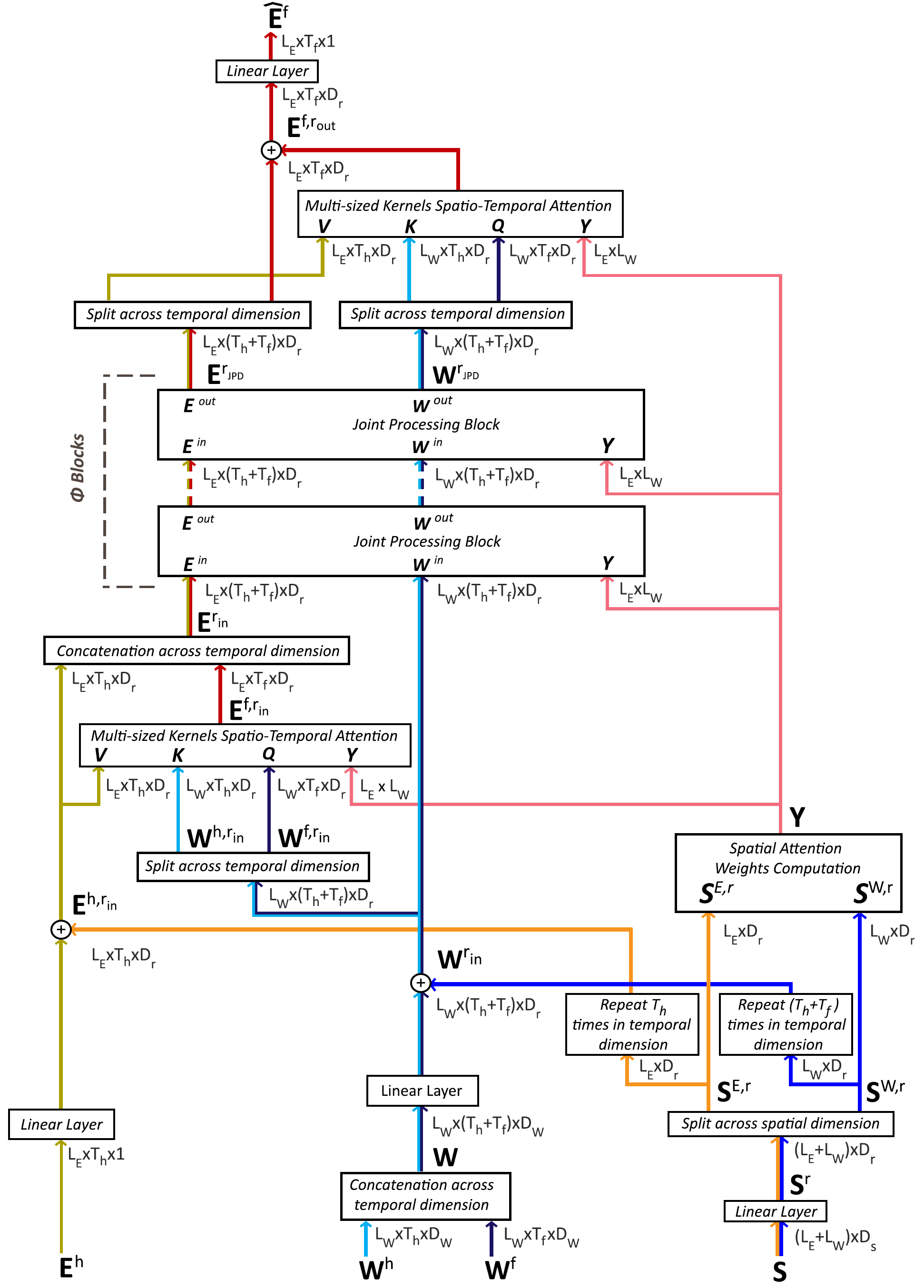}
    \caption{Architecture of the proposed wind/solar energy forecasting method.}
    \label{fig:architecture}
\end{figure}

The overall architecture of the proposed method is depicted on Figure~\ref{fig:architecture}. 
The method receives as input $\textbf{E}^{h}$, $\textbf{W}^{h}$ and $\textbf{W}^{f}$ and feeds them into respective linear projection layers, forming  $\textbf{E}^{h, r_{in}}  \in \mathbb{R}^{L_E \times T_h \times D_r}$, $\textbf{W}^{h, r_{in}} \in \mathbb{R}^{L_W \times T_h \times D_r}$ and $\textbf{W}^{f,r_{in}}  \in \mathbb{R}^{L_W \times T_f \times D_r}$, respectively. In this formulation,  $L_E$ is the number of energy generation sites, $L_W$ is the number of the weather data sites, $T_h$ is the number of past time-steps. In addition, the method  receives spatial encodings $\textbf{S} \in \mathbb{R}^{(L_E +L_W)\times D_S}$, where $D_S$ is the size of the encodings. Although these encodings can be generated based on the geolocation coordinates of the respective sites, if these are not known, which is the case studied in this paper, one can use the one-hot encoding scheme for generating discrete pseudo-spatial encodings for all sites. The encodings are fed to a linear layer, in order to linearly project them into a $D_r$-dimensional space. Since the method processes energy generation data from all sites using the same neural layers, the projected pseudo-spatial encodings of the energy generation sites are added to $\textbf{E}^{h, r_{in}}$, so as to inject the spatial information into the neural network. Similarly, the pseudo-spatial encodings of weather data are added to $\textbf{W}^{h, r_{in}}$ and  $\textbf{W}^{f, r_{in}}$. In addition, those spatial encodings are utilized for the generation of matrix $\textbf{Y} \in \mathbb{R}^{L_E \times L_W}$, which holds the spatial relations between the $L_E$ energy generation locations and the $L_W$  weather data locations. The spatial attention weights matrix can be defined as:

\begin{equation}
    \textbf{Y} =  softmax(\frac{\boldsymbol{\Lambda}^{E}  \boldsymbol{\Lambda}^{W\text{ }T}}{\sqrt{D_\Lambda}})
\end{equation}
\begin{equation}
\text{where \ } \boldsymbol{\Lambda}^{E} = \textbf{S}^{E,r} \textbf{C}^{E} \text{ \ and \ } \boldsymbol{\Lambda}^{W} = \textbf{S}^{W,r} \textbf{C}^{W}
\end{equation}

\noindent In this formulation, the attention weights in \textbf{Y} are computed similar to Equation~\ref{eqn:atten_scores}, through $\boldsymbol{\Lambda}^{E} \in \mathbb{R}^{L_E \times D_\Lambda}$ and $\boldsymbol{\Lambda}^{W} \in \mathbb{R}^{L_W \times D_\Lambda}$ matching. $\textbf{S}^{E, r} \in \mathbb{R}^{L_E \times D_r}$ and $\textbf{S}^{W, r} \in \mathbb{R}^{L_W \times D_r}$ are the projected spatial encodings of energy generation sites and weather data sites, respectively. In addition, $\textbf{C}^{E}  \in \mathbb{R}^{D_r \times D_\Lambda}$ and $\textbf{C}^{W}  \in \mathbb{R}^{D_r \times D_\Lambda}$ are learnable parameters of the linear projection layers.

Aiming to generate future-time energy generation latent representations $\textbf{E}^{f, r_{in}}  \in \mathbb{R}^{L_E \times T_f \times D_r}$ we employ the MKST-Attention mechanism described in Section~\ref{sec:mksta} as follows:

\begin{equation}
\textbf{E}^{f, r_{in}} = MKST\text{-}Attention(\textbf{W}^{f, r_{in}},\textbf{W}^{h, r_{in}}, \textbf{E}^{h, r_{in}}, \textbf{Y})
\end{equation}

Then, $\textbf{E}^{r_{in}} \in \mathbb{R}^{L_E \times (T_h+T_f) \times D_r}$, which results from the  concatenation of $\textbf{E}^{h, r_{in}}$ and $\textbf{E}^{f, r_{in}}$ in the temporal dimension and $\textbf{W}^{r_{in}}  \in \mathbb{R}^{L_W \times (T_h+T_f) \times D_r}$, which results from the  concatenation of $\textbf{W}^{h, r_{in}}$ and $\textbf{W}^{f, r_{in}}$ also in the temporal dimension, are sequentially processed by $\phi$  stacked Joint Processing Blocks, resulting in refined latent energy generation and weather data representations, namely $\textbf{E}^{r_{JPB}} \in \mathbb{R}^{L_E \times (T_h+T_f) \times D_r}$  and $\textbf{W}^{r_{JPB}} \in \mathbb{R}^{L_W \times (T_h+T_f) \times D_r}$ ($r_{JPB}$: representations derived through Joint Processing Blocks). Thereafter, the final representation of energy generation predictions $\textbf{E}^{f, r_{out}}  \in \mathbb{R}^{L_E \times T_f \times D_r}$ are computed as follows:

\begin{equation}
\textbf{E}^{f, r_{out}} = MKST\text{-}Attention(\textbf{W}^{f, r_{JPB}}, \textbf{W}^{h, r_{JPB}}, \textbf{E}^{h, r_{JPB}}, \textbf{Y}) +  \textbf{E}^{f, r_{JPB}}
\end{equation}

\noindent where $\textbf{E}^{h, r_{JPB}} \in \mathbb{R}^{L_E \times T_h \times D_r}$ and $\textbf{E}^{f, r_{JPB}} \in \mathbb{R}^{L_E \times T_f \times D_r}$ are past- and future-time energy generation representations and $\textbf{W}^{h, r_{JPB}} \in \mathbb{R}^{L_W \times T_h \times D_r}$,  $\textbf{W}^{f, r_{JPB}} \in \mathbb{R}^{L_W \times T_f \times D_r}$ are past- and future-time weather data representations, all generated through the $\phi$ stacked Joint Processing Blocks. Finally, $\textbf{E}^{f, r_{out}}$ is fed into a linear layer, thus obtaining the energy generation predictions $\hat{\textbf{E}}^{f}$.

\section{Experimental Evaluation}\label{exp_eval}

 The experimental evaluation of the proposed method was conducted in a day-ahead (forecasting horizon of 24 hours) solar/wind energy generation forecasting scenario, utilizing weather data (predictions) for the corresponding forecasting window, as well as past energy generation data and weather data for the last 14 days (past-time measurements of 336 hours). Specifically for the solar energy forecasting task, time-related features, such as the hour of the day, the month of the year and the season, were treated as cyclical features and were encoded in polar coordinates. Then, they were concatenated along with weather data before being fed to the proposed method. It shall be noted that the time-related features were not employed in the wind energy forecasting task due to the lack of a noticeable correlation between them and the wind energy generation.

The rest of the section is structured as follows. First, details of the utilized datasets are provided, followed by the description  of the proposed method implementation details.
Next, the selected baseline methods are described and, finally, the experimental results are provided and discussed.

\subsection{Dataset Description}
Five datasets were used in the experiments, as detailed below.\\
\textbf{GEFCom2014 wind/solar}: The complete dataset, named \textit{GEFCom2014 dataset}  \cite{hong2016gefcom}, contains four parts, each for a specific energy-related forecasting task. These tasks are: (i) electric load forecasting, (ii) electricity price forecasting, (iii) wind energy forecasting and (iv) solar energy forecasting. In our experimental evaluation, only the latter two were employed. According to the authors in \cite{hong2016gefcom}, \textit{GEFCom2014-wind} contains wind energy data and wind-related weather data obtained from ten power stations in Australia, featuring an hourly temporal resolution. The locations of the stations are not disclosed. Both the energy data and the wind-related weather data span the period from  2012 to 2013. The weather data contain the zonal component $u$ and meridional component $v$ of the wind velocity vector estimated at 10m and 100m above ground level. In a similar manner, \textit{GEFCom2014-solar} contains solar energy generation data as well as weather data consisting of 12 variables, such as Total Cloud Cover (TCC), Surface Solar Rad Down (SSRD), Surface Thermal Rad Down (STRD), Total Column Liquid Water (TCLW) and Surface Pressure (SP), from three solar power stations in Australia, whose exact locations are not provided. Both the energy data and the wind-related weather data span the period from 2012 to 2014. The weather data of both sub-datasets were obtained from the European Centre for Medium-range Weather Forecasts (ECMWF) and were to be issued each day at midnight. The solar energy data, along with its associated weather data, maintains a temporal resolution of one hour. Both \textit{GEFCom2014-wind} and \textit{GEFCom2014-solar} were proposed for day-ahead probabilistic energy prediction, with an hourly resolution. However, in our experiments we employed them  for deterministic day-ahead wind and solar energy generation forecasting at an hour-level temporal resolution. Details regarding the dataset's split are provided in Table~\ref{Tab:gefcom_split}.

\begin{table}[!htb]
    \begin{tabular}{|c|c|c|}
        \hline
        \multirow{ 2}{*}{\textbf{Set}} & \multicolumn{2}{c|}{\textbf{Forecasting Period 
        }} \\ \hhline{~--}
         & \textbf{\textit{GEFCom2014-wind}} & \textbf{\textit{GEFCom2014-solar}}\\ \hline
          \quad Training set \quad & \quad  15/01/2012 \quad - \quad 31/10/2013 \quad & \quad  16/04/2012 \quad - \quad 30/04/2014 \quad \\ \hline
          \quad Validation set \quad & \quad 01/11/2013 \quad - \quad 30/11/2013 \quad & \quad  01/05/2014 \quad - \quad 31/05/2014 \quad \\ \hline
          \quad Testing set \quad & \quad 01/12/2013 \quad - \quad 30/12/2013 \quad & \quad 01/06/2014 \quad - \quad 30/06/2014 \quad \\ \hline
    \end{tabular} 
    \caption{Training\slash Validation\slash Testing split of \textit{GEFCom2014-wind/solar}}
    \label{Tab:gefcom_split}
\end{table}

\textbf{AEMO-H}: The dataset was originally proposed in \cite{dowell2016aemo} for very short-term probabilistic wind energy forecasting. It contains 5-minute resolution wind energy data from 22 wind farms in south-east Australia from the period 2012-2013 and it was made publicly available by the Australian Electricity Market Operator (AEMO). In our  experimental evaluation, two modifications / enhancements were applied. First, 5-minute resolution weather data, corresponding to the locations of the  wind farms, were retrieved from the Solcast\footnote{\url{https://solcast.com/}} online service. Details regarding the retrieved weather data are presented in Table~\ref{Tab:aemo_nwp}. Second, the dataset's temporal resolution was  modified from 5 minutes to 1 hour, aiming to employ the dataset in an 1-hour resolution, day-ahead wind energy generation forecasting scenario. The  temporal resolution was modified by  averaging the  wind energy generation values as well as the average  of weather data variables within each hour interval. Details about the dataset's split are presented in Table~\ref{Tab:aemo_split}.

\begin{table}[!htb]
    \begin{subtable}{.5\textwidth}
        \begin{tabular}{|c|c|c|c|}
            \hline
            \multicolumn{4}{|c|}{\textbf{Weather Data Variables}} \\ \hline
            \multicolumn{2}{|c}{Air temperature $(\tccentigrade)$} & \multicolumn{2}{|c|}{Dew point temperature $(\tccentigrade)$}  \\ \hline
            \multicolumn{2}{|c}{Precipitable water $(kg/m^2)$} & \multicolumn{2}{|c|}{Precipitable rate $(mm/h)$} \\ \hline
            \multicolumn{2}{|c|}{ Wind speed 10m/100m $(m/s)$} &  \multicolumn{2}{c|}{ Wind direction 10m/100m $(\degree)$} \\ \hline
            \multicolumn{4}{|c|}{ Surface pressure $(hPa)$} \\ \hline

             \end{tabular}
        \caption{Weather data variables}
        \label{Tab:aemo_nwp}
    \end{subtable} 
    \begin{subtable}{.3\textwidth}
        \begin{tabular}{|c|c|}
            \hline
            \textbf{Set} & \textbf{Forecasting Period} \\ \hline
             Training set & 15/01/2012 - 30/11/2012 \\ \hline
             Validation set &  01/12/2012 - 31/12/2012 \\ \hline
             Testing set &  01/01/2013 - 30/12/2013 \\ \hline
        \end{tabular} 
        \caption{Training\slash Validation\slash Testing split}         \label{Tab:aemo_split}
    \end{subtable}
    \caption{Details regarding AEMO-H.}
\end{table}

\textbf{ENTSO-E wind/solar}: The datasets consist of (i) hourly wind/solar energy generation data for Greece (the entire country), collected by the European Network of Transmission System Operators for Electricity\footnote{\url{https://transparency.entsoe.eu}} (ENTSO-E), and (ii) hourly weather data for 29 locations in Greece, retrieved by the OpenWeather\footnote{\url{https://openweathermap.org/}} online service. The coordinates of these locations are depicted in Table~\ref{Tab:entsoe_locs}. These locations coincide with locations of large solar and wind power stations in Greece. The  weather data variables that were employed in the experiments are listed in Table~\ref{Tab:entsoe_nwp}. The dataset spans the period 2017-2023. Details about the dataset's split are depicted in Table~\ref{Tab:entsoe_split}. Preliminary or modified versions of the datasets were originally employed in \cite{symeonidis2023wind} and \cite{vartholomaios2021short}.

\begin{table}[!htb]
        \resizebox{0.8\columnwidth}{!}{
        \begin{subtable}{\textwidth}
        \begin{tabular}{|c|c|c||c|c|c||c|c|c|}
            \hline
            \textbf{ID} & \textbf{Lat.} & \textbf{Lon.} &  \textbf{ID} & \textbf{Lat.} & \textbf{Lon.} &  \textbf{ID} & \textbf{Lat.} & \textbf{Lon.}\\ \hline
            1 & 37.5411 & 22.5951 & 11 & 41.1655 & 25.9326 & 21 & 38.2119 & 23.4278 \\ \hline
            
            2 & 37.7662 & 22.8489 & 12 & 41.1528 & 25.8356 & 22 & 41.2958  & 25.8879 \\ \hline
            
            3 & 41.0472 & 25.9823 & 13 & 38.1886 & 23.2295 & 23 & 40.4103 & 22.0868 \\ \hline
            
            4 & 37.4915 & 23.1672 & 14 & 38.4691 & 23.2353 & 24 & 41.1553 & 25.9094 \\ \hline
            
            5 & 37.4313 & 22.3016 & 15 & 38.4066 & 21.8195 & 25 & 41.2986 & 23.3747 \\ \hline
            
            6 & 37.5007 & 23.3038 & 16 & 36.8331 & 22.9094 & 26 & 38.2305 & 23.5026 \\ \hline
            
            7 & 38.1395 & 22.1279 & 17 & 38.2889 & 20.4995 & 27 & 38.3603 & 23.2384 \\ \hline
            
            8 & 40.8091 & 21.2575 & 18 & 41.2038 & 24.2845 & 28 & 38.4308 & 22.2748 \\ \hline
            
            9 & 37.4779 & 23.2937 & 19 & 40.9651 & 25.9702 & 29 & 36.8554 & 22.9517 \\ \hline
            
            10 & 37.5286 & 23.2460 & 20 & 38.2226 & 31.8676 & - & - & - \\ \hline
             \end{tabular}
        \caption{Coordinates of the 29 weather locations.}
        \label{Tab:entsoe_locs}
    \end{subtable} 
    }
    \begin{subtable}{.5\textwidth}
        \begin{tabular}{|c|c|c|}
            \hline
            \multicolumn{3}{|c|}{\textbf{Weather Data Variables}} \\ \hline
            \multicolumn{3}{|c|}{Min\slash max\slash feels like\slash actual temperature $(\tccentigrade)$} \\ \hline
            \multicolumn{2}{|c|}{Atmospheric Pressure $(hPa)$} & Humidity (\%)\\ \hline
            Cloudiness (\%) & \multicolumn{1}{c|}{ Wind direction (\%)} & Wind speed $(m/s)$\\ \hline
             \end{tabular}
        \caption{Types of NWPs}
        \label{Tab:entsoe_nwp}
    \end{subtable} 
    \begin{subtable}{.5\textwidth}
        \begin{tabular}{|c|c|}
            \hline
            \textbf{Set} & \textbf{\quad \quad Forecasting Period \quad \quad } \\ \hline
             \quad Training set \quad & \quad 15/01/2017 - 31/12/2020 \quad \\ \hline
             \quad Validation set \quad & \quad 01/01/2021 - 31/12/2021 \quad \\ \hline
             \quad Testing set \quad & \quad 01/01/2022 - 30/01/2023 \quad \\ \hline
        \end{tabular} 
        \caption{Training\slash Validation\slash Testing split}         \label{Tab:entsoe_split}
    \end{subtable}
    \caption{Details regarding ENTSO-E wind/solar.}
\end{table}

\subsection{Implementation details}

The majority of parameters related to the proposed method  architecture were set universally for all datasets. More specifically, the size of latent representations $D_r$, for both energy generation data and weather data, was set to 48. In MKST-Attention, $D_K$ and $D_V$ were set to 16. The selected kernel sizes were set to [3, 5, 7]. Thus, $\Upsilon$, namely the number of the kernel sizes, was set to 3. Regarding UTCAE, the number of pyramid levels $P$ was set to 4. Finally, the number  $\phi$ of JPB blocks was set to 3. 
To prevent over-fitting, dropout regularization was implemented with a rate between 0.1 and 0.2. In addition, Reversible Instance Normalization~\cite{kim2021reversible} (RevIN) was implemented for addressing the distribution shift problem.

The method was trained by utilizing the Mean Squared Error (MSE) as loss function:

\begin{equation}
    \label{Eq:loss}
    \mathcal{L} = \frac{1}{L_E}\frac{1}{T_f}\sum_{s=1}^{L_E}  \sum_{t=1}^{T_f} (\hat{e} \langle s,t \rangle - e\langle s,t \rangle)^2  
\end{equation}
The training parameters were also set universally for all datasets. Adam optimizer was used during training, with the initial learning rate set to $5 \times 10^{-4}$. The ReduceLROnPlateau\footnote{\url{https://pytorch.org/docs/stable/generated/torch.optim.lr_scheduler.ReduceLROnPlateau.html}} scheduler, was employed for reducing the learning accordingly. The early stopping technique was employed for finding the optimal number of training epochs, based on the performance of the method on the validation set of each dataset. The implemented model, which supports batch-training, was trained using batches of 8 samples.

\subsection{Baseline Methods}

Nine neural and two non-neural SoA time-series forecasting methods were selected as baselines for comparison against the proposed method in the wind/solar energy forecasting tasks. All methods, except \cite{symeonidis2023wind}, are time-series forecasting methods that haven't been explicitly proposed for wind/solar energy forecasting. The rationale for this selection stemmed from the fact that none of the wind/solar energy forecasting methods outlined in Section~\ref{Sec:rel_work} is provided in open source. It shall be noted however that the majority of these papers follow a similar experimental evaluation methodology wherein the respective proposed method is compared against generic time-series forecasting methods.

Table~\ref{Tab:results_methods} lists all baseline methods, and indicates the type (past-time, future-time, or both) of covariates supported by each method. In all methods, past energy generation measurements were employed as lagged input variables, while weather data were employed as covariates. Specifically for the solar energy forecasting task,  we utilized as covariates the concatenation of weather data and time-related features.

Implementations of all baseline methods, except \cite{symeonidis2023wind}, were retrieved by the \textit{Darts}\cite{herzen2022darts} library. Thus, details about any referred implementation modifications against the originally proposed version of each baseline method, can be found in Darts official GitHub repository\footnote{\url{https://github.com/unit8co/darts}}. The Asynchronous Successive Halving Algorithm~\cite{asha2020} (ASHA) scheduler, was utilized for hyperparameter tuning and for selecting  the initial learning rate. Similar to our method, the ReduceLROnPlateau scheduler was employed for reducing the learning accordingly and the early stopping technique was employed for finding the optimal number of training epochs, based on the performance of each method on the validation set of each dataset. In all methods, MSE was used as the corresponding loss function. Finally, all baseline methods, except \cite{symeonidis2023wind}, were employed in a single-site manner. In other words, a separate model was trained and evaluated   for each energy generation site, and only  weather data for the corresponding site were used. In ENTSO-E wind and solar datasets, the weather data were spatially aggregated before being fed as covariates, instead of being fed separately for each of the 29 locations. This choice was made based on preliminary experiments, where the baseline methods, using weather data from all 29 locations, achieved better results in training set but performed poorly in validation and testing sets, demonstrating the inability of the methods to select weather data from relevant locations, leading thus to over-fitting.

The two selected non-neural time-series forecasting methods, both  utilizing future-time covariates as input, are Random Forest regression~\cite{breiman2001random} and Kalman Filtering~\cite{morrison1977kalman}. Among the selected neural time-series forecasting methods, \textit{N-HiTS}~\cite{nhits_2023} combines two complementary techniques, multi-rate input sampling and hierarchical interpolation, to produce drastically improved, interpretable, and computationally efficient long-horizon time-series predictions. N-HiTS is a univariate method, accepting as input only the predicted time-series history. In the conducted evaluation, a modified version of N-HiTS, presented in \cite{herzen2022darts}, was employed, supporting the use of past-time weather data as past-time covariates. In addition, an implementation of a dilated Temporal Convolutional Neural Network (TCNN), suitable for time-series forecasting  inspired from \cite{tcn2018}, was employed. The utilized implementation also supports the use of past-time covariates. An implementation of the Transformer~\cite{transformer2017} model, supporting past-time weather-data as covariates, was also included in the conducted  evaluation. Furthermore, a deterministic autoregressive recurrent neural network, inspired by \textit{DeepAR}\cite{deepar2020}, was  employed for solar/wind energy generation forecasting. The network is capable of efficiently receiving and processing future-time covariates as input. DLinear and NLinear are two simple one-layer linear models, proposed in \cite{dnlinear2023} for long-term time-series forecasting tasks. Both  utilize past- and future-time covariates as input. Moreover, Time-series Dense Encoder (TiDE)~\cite{tide2023} is a Multi-layer Perceptron (MLP)-based encoder-decoder model, suitable for long-term time-series forecasting tasks, that enjoys the simplicity and speed of linear models while also being able to handle past- and future-time covariates and non linear dependencies. Additionally, the Temporal Fusion Transformer (TFT)~\cite{tft2021} method adopts an attention-based architecture that is able to learn temporal relationships at different scales, using recurrent layers for local processing and interpretable self-attention layers for long-term dependencies. TFT allows the use of both past- and future-time covariates, while utilizing specialized components to select relevant features and a series of gating layers to suppress unnecessary components. Finally, the method in \cite{symeonidis2023wind}, designed specifically for wind energy forecasting based on weather data from multiple locations, is also employed in the experimental evaluation process on both wind and solar energy forecasting tasks. As already mentioned in Section~\ref{introduction}, this  method is a preliminary version of the one proposed in this paper. The key differences between the two methods are highlighted in the same Section.

\begin{center}
\begin{table}[ht]
\resizebox{0.6\columnwidth}{!}{

\begin{tabular}{|l|l|l|}
\hline
\multirow{3}{*}{\textbf{Method}} &
 \multicolumn{2}{c|}{\textbf{Covariates}}
\\ \hhline{~--}
& 
\multirow{2}{*}{\pbox[c][0.6cm][c]{22cm}{\textbf{Past-time}}} & \multirow{2}{*}{\pbox[c][0.6cm][c]{22cm}{\textbf{Future-time}}}
\\
& & \\ \hline
\multicolumn{1}{|l|}{N-HiTS~\cite{nhits_2023}} & \multicolumn{1}{c|}{\checkmark} & \multicolumn{1}{c|}{} \\ \hline

\multicolumn{1}{|l|}{TCNN~\cite{tcn2018}} & \multicolumn{1}{c|}{\checkmark} & \multicolumn{1}{c|}{} \\ \hline

\multicolumn{1}{|l|}{Transformer~\cite{transformer2017}} & \multicolumn{1}{c|}{\checkmark} & \multicolumn{1}{c|}{} \\ \hline

\multicolumn{1}{|l|}{Kalman Filtering~\cite{morrison1977kalman}} & \multicolumn{1}{c|}{} & \multicolumn{1}{c|}{\checkmark} \\ \hline

\multicolumn{1}{|l|}{Random Forest~\cite{breiman2001random}} & \multicolumn{1}{c|}{} & \multicolumn{1}{c|}{\checkmark} \\ \hline

\multicolumn{1}{|l|}{DeepAR-based RNN~\cite{deepar2020}} & \multicolumn{1}{c|}{} & \multicolumn{1}{c|}{\checkmark} \\ \hline

\multicolumn{1}{|l|}{DLinear~\cite{dnlinear2023}} & \multicolumn{1}{c|}{\checkmark} & \multicolumn{1}{c|}{\checkmark} \\ \hline

\multicolumn{1}{|l|}{NLinear~\cite{dnlinear2023}} & \multicolumn{1}{c|}{\checkmark} & \multicolumn{1}{c|}{\checkmark} \\ \hline

\multicolumn{1}{|l|}{TiDE~\cite{tide2023}} & \multicolumn{1}{c|}{\checkmark} & \multicolumn{1}{c|}{\checkmark} \\ \hline

\multicolumn{1}{|l|}{TFT~\cite{tft2021}} & \multicolumn{1}{c|}{\checkmark} & \multicolumn{1}{c|}{\checkmark} \\ \hline

\multicolumn{1}{|l|}{\cite{symeonidis2023wind}} & \multicolumn{1}{c|}{\checkmark} & \multicolumn{1}{c|}{\checkmark} \\ \hline

\multicolumn{1}{|l|}{Ours} & \multicolumn{1}{c|}{\checkmark} & \multicolumn{1}{c|}{\checkmark}  \\ \hline    
\end{tabular}
}
\caption{\centering Comparison of the employed time-series forecasting methods in deterministic RES forecasting.}
\label{Tab:results_methods}
\end{table}
\end{center}

\subsection{Results}

Mean Absolute Error (MAE) and Root Mean Square Error (RMSE) were utilized to measure the performance of the methods in each energy generation site. To achieve a fair comparison, each method was trained four times from scratch with the same setup, and the mean MAE and RMSE are reported. For each dataset, both metrics are reported separately for each energy generation site. The average MAE and RMSE values, computed across all sites, are also provided. 

\textbf{GEFCom2014-wind:} Tables~\ref{Tab:gefcom_w_mae} and \ref{Tab:gefcom_w_rmse} report MAE and RMSE values achieved by the employed methods, respectively. Our proposed method attained the best average results, computed across the ten wind energy generation sites, in both evaluation metrics. More specifically it achieved MAE and RMSE values of 0.108 and 0.160 respectively. Among the rest of the methods, TFT, the DeepAR-based RNN and the method proposed in \cite{symeonidis2023wind}, which, as already mentioned, is a preliminary version of the one proposed in this paper, achieved favourable performance results. In particular, TFT achieved top MAE, being on par with our proposed method, in sites III and IV. In addition, TFT attained top RMSE in sites IV, V. The method proposed in \cite{symeonidis2023wind} achieved top RMSE in site III.

\begin{table}[ht]
\centering
\resizebox{\columnwidth}{!}{
\begin{tabular}{|c|c|c|c|c|c|c|c|c|c|c||c|}
\hline
\multirow{2}{*}{\textbf{Method}} & \multicolumn{11}{c|}{\textbf{Energy Stations}} \\ \hhline{~-----------}

& \textbf{I} & \textbf{II} & \textbf{III} & \textbf{IV} & \textbf{V} & \textbf{VI}
& \textbf{VII} & \textbf{VIII} & \textbf{IX} & \textbf{X} & \textbf{All}  \\ \hline

TCNN &
0.206 & 0.180 & 0.231 & 0.252 & 0.263 & 0.263 & 0.203 & 0.201 & 0.190 & 0.265 & 0.225\\ \hline
N-HiTS & 
0.188 & 0.185 & 0.205 & 0.227 & 0.225 & 0.225 & 0.182 & 0.183 & 0.185 & 0.243 & 0.205\\ \hline
Transformer & 
0.181 & 0.166 & 0.203 & 0.211 & 0.216 & 0.229 & 0.168 & 0.176 & 0.182 & 0.223 & 0.195\\ \hline
Kalman & 
\multirow{2}{*}{0.167} & \multirow{2}{*}{0.144} & \multirow{2}{*}{0.193} &
\multirow{2}{*}{0.172} & \multirow{2}{*}{0.195} & \multirow{2}{*}{0.202} & 
\multirow{2}{*}{0.162} & \multirow{2}{*}{0.170} & \multirow{2}{*}{0.154} &
\multirow{2}{*}{0.207} & \multirow{2}{*}{0.177} \\ 
Filtering & 
& & & & & & & & & & \\ \hline
Random & 
\multirow{2}{*}{0.152} & \multirow{2}{*}{0.151} & \multirow{2}{*}{0.187} &
\multirow{2}{*}{0.178} & \multirow{2}{*}{0.181} & \multirow{2}{*}{0.194} &
\multirow{2}{*}{0.138} & \multirow{2}{*}{0.155} & \multirow{2}{*}{0.136} &
\multirow{2}{*}{0.215} & \multirow{2}{*}{0.169} \\ 
Forest & 
& & & & & & & & & & \\ \hline
DeepAR-based & 
\multirow{2}{*}{0.125} & 
\multirow{2}{*}{0.134} &
\multirow{2}{*}{0.121} &
\multirow{2}{*}{0.126} &
\multirow{2}{*}{0.141} &
\multirow{2}{*}{0.151} &
\multirow{2}{*}{0.082} & 
\multirow{2}{*}{0.105} & 
\multirow{2}{*}{0.110} &
\multirow{2}{*}{0.164} &
\multirow{2}{*}{0.126} \\
RNN & & & & & & & & & & & \\ \hline

DLinear & 
0.177 & 0.181 & 0.208 & 0.216 & 0.227 & 0.224 & 0.177 & 0.176 & 0.170 & 0.250 & 0.201\\ \hline
NLinar & 
0.201 & 0.188 & 0.211 & 0.209 & 0.222 & 0.226 & 0.204 & 0.205 & 0.194 & 0.259 & 0.212\\ \hline
TiDE & 
0.135 & 0.122 & 0.130 & 0.138 & 0.152 & 0.156 & 0.084 & 0.107 & 0.107 & 0.185 & 0.132 \\ \hline
TFT & 
0.111 & 0.110 & \textbf{0.121} & \textbf{0.118} & 0.134 & 0.141 & 0.074 & 0.092 & 0.093 & 0.159 & 0.115 \\ \hline 
\cite{symeonidis2023wind} & 
0.099 & 0.115 & 0.123 & 0.140 & 0.152 & 0.150 & 0.084 & 0.099 & 0.100 & 0.164 & 0.122 \\ \hline 
\hline
Ours & 
\textbf{0.077} & \textbf{0.099} & \textbf{0.121} & \textbf{0.118} & \textbf{0.132} & \textbf{0.133} & \textbf{0.070} & \textbf{0.090} & \textbf{0.090} & \textbf{0.150} & \textbf{0.108} \\ \hline
\end{tabular}
}
\caption{Mean Absolute Error (MAE) of the compared methods in GEFCom2014-wind dataset.}
\label{Tab:gefcom_w_mae}
\end{table}

\begin{table}[ht]
\centering
\resizebox{\columnwidth}{!}{
\begin{tabular}{|c|c|c|c|c|c|c|c|c|c|c||c|}
\hline
\multirow{2}{*}{\textbf{Method}} & \multicolumn{11}{c|}{\textbf{Energy Stations}} \\ \hhline{~-----------}

& \textbf{I} & \textbf{II} & \textbf{III} & \textbf{IV} & \textbf{V} & \textbf{VI}
& \textbf{VII} & \textbf{VIII} & \textbf{IX} & \textbf{X} & \textbf{All}  \\ \hline

TCNN & 
0.267 & 0.222 & 0.277 & 0.303 & 0.305 & 0.310 & 0.250 & 0.257 & 0.238 & 0.317 & 0.275\\ \hline
N-HiTS & 
0.254 & 0.233 & 0.251 & 0.275 & 0.268 & 0.269 & 0.232 & 0.230 & 0.228 & 0.296 & 0.254\\ \hline
Transformer & 
0.245 & 0.212 & 0.249 & 0.261 & 0.278 & 0.292 & 0.213 & 0.225 & 0.225 & 0.284 & 0.248\\ \hline
Kalman & 
\multirow{2}{*}{0.223} & \multirow{2}{*}{0.189} & \multirow{2}{*}{0.237} &
\multirow{2}{*}{0.215} & \multirow{2}{*}{0.250} & \multirow{2}{*}{0.256} &
\multirow{2}{*}{0.204} & \multirow{2}{*}{0.214} & \multirow{2}{*}{0.198} & 
\multirow{2}{*}{0.254} & \multirow{2}{*}{0.224}\\
Filtering & & & & & & & & & & & \\ \hline
Random & 
\multirow{2}{*}{0.206} & \multirow{2}{*}{0.195} & \multirow{2}{*}{0.238} &
\multirow{2}{*}{0.226} & \multirow{2}{*}{0.228} & \multirow{2}{*}{0.239} &
\multirow{2}{*}{0.176} & \multirow{2}{*}{0.201} & \multirow{2}{*}{0.180} & 
\multirow{2}{*}{0.266} & \multirow{2}{*}{0.215} \\
Forest & & & & & & & & & & & \\ \hline
DeepAR-based & 
\multirow{2}{*}{0.176} &
\multirow{2}{*}{0.187} &
\multirow{2}{*}{0.162} & 
\multirow{2}{*}{0.183} &
\multirow{2}{*}{0.197} &
\multirow{2}{*}{0.221} & 
\multirow{2}{*}{0.115} & 
\multirow{2}{*}{0.157} &
\multirow{2}{*}{0.151} & 
\multirow{2}{*}{0.234} & 
\multirow{2}{*}{0.178}\\
RNN & & & & & & & & & & & \\ \hline
DLinear & 
0.245 & 0.230 & 0.256 & 0.263 & 0.271 & 0.272 & 0.227 & 0.228 & 0.217 & 0.302 & 0.251\\ \hline
NLinar & 
0.254 & 0.239 & 0.252 & 0.258 & 0.266 & 0.270 & 0.237 & 0.237 & 0.230 & 0.311 & 0.255\\ \hline
TiDE & 
0.183 & 0.166 & 0.173 & 0.182 & 0.201 & 0.212 & 0.115 & 0.160 & 0.149 & 0.254 & 0.180\\ \hline
TFT & 
0.158 & 0.159 & 0.166 & \textbf{0.179} & \textbf{0.191} & 0.208 &
0.105 & 0.147 & 0.134 & 0.230 & 0.168 \\ \hline
\cite{symeonidis2023wind} & 
0.132 & 0.161 & \textbf{0.161} & 0.193 & 0.202 & 0.202 & 0.111 & 0.148 & 0.137 & 0.225 & 0.167 \\ \hline\hline
Ours & 
\textbf{0.110} & \textbf{0.148} & 0.163 & 0.180 & 0.198 & \textbf{0.201} & \textbf{0.097} & \textbf{0.144} & \textbf{0.133} & \textbf{0.224} & \textbf{0.160}\\ \hline

\end{tabular}
}
\caption{Root Mean Square Error (RMSE) of the compared methods in GEFCom2014-wind dataset.}
\label{Tab:gefcom_w_rmse}
\end{table}

\textbf{GEFCom2014-solar:} Table~\ref{Tab:gefcom_s_mae_rmse} reports the achieved MAE and RMSE for the proposed and the baseline methods in GEFCom2014-solar dataset. Our  method attained the best-average results, across the three solar energy generation sites, in both evaluation metrics. More specifically it achieved MAE and RMSE values of 0.0362 and 0.0792 respectively. Similar to GEFCom2014-wind dataset, TFT, the DeepAR-based RNN and the method in \cite{symeonidis2023wind} achieved good results in terms of MAE and RMSE. In particular, the DeepAR-based RNN managed to achieve the lowest MAE of 0.0386 in station II. In the same station, the method proposed in \cite{symeonidis2023wind} achieved top RMSE of 0.830, being on par with our proposed method.

\begin{table}[ht]
\vspace{-0.5cm}
\centering
\caption{Mean Absolute Error (MAE) and Root Mean Square Error (RMSE) of the compared methods in GEFCom2014-solar dataset.}
\label{Tab:gefcom_s_mae_rmse}
\begin{tabular}{c}
\resizebox{0.97\columnwidth}{!}{
\begin{tabular}{|c|cc|cc|cc||cc|}
\hline
\multirow{3}{*}{\textbf{Method}} & \multicolumn{8}{c|}{\textbf{Energy Stations}} \\ \hhline{~--------}

& \multicolumn{2}{c|}{\textbf{I}} & \multicolumn{2}{c|}{\textbf{II}} & \multicolumn{2}{c||}{\textbf{III}} & \multicolumn{2}{c|}{\textbf{All}}  \\

 & \textbf{MAE} & \textbf{RMSE} & \textbf{MAE} & \textbf{RMSE} & \textbf{MAE} & \textbf{RMSE} & \textbf{MAE} & \textbf{RMSE}  \\ \hline

TCNN & 
0.0506 & 0.1036 & 0.0514 & 0.0990 & 0.0532 & 0.1004 & 0.0517 & 0.1010 \\ \hline
N-HiTS & 
0.0534 & 0.1038 & 0.0517 & 0.0956 & 0.0555 & 0.1013 & 0.0535 & 0.1002 \\ \hline
Transformer & 
0.0571 & 0.1091 & 0.0557 & 0.1036 & 0.0566 & 0.01027 & 0.0564 & 0.1051 \\ \hline
Kalman & 
\multirow{2}{*}{0.0635}	& \multirow{2}{*}{0.0994} & \multirow{2}{*}{0.0737} & \multirow{2}{*}{0.1092} & \multirow{2}{*}{0.0617} & \multirow{2}{*}{0.0998} &
\multirow{2}{*}{0.0663} & \multirow{2}{*}{0.1028} \\ 
Filtering & & & & & & &  & \\ \hline
Random & 
\multirow{2}{*}{0.0571}	& \multirow{2}{*}{0.1091} & \multirow{2}{*}{0.0557} &
\multirow{2}{*}{0.1036} & \multirow{2}{*}{0.0566} & \multirow{2}{*}{0.1027} &
\multirow{2}{*}{0.0564} & \multirow{2}{*}{0.1051} \\ 
Forest & & & & & & &  & \\ \hline
DeepAR-based  & 
\multirow{2}{*}{0.0348} & \multirow{2}{*}{0.0825} & \multirow{2}{*}{\textbf{0.0386}} & \multirow{2}{*}{0.0871} & \multirow{2}{*}{0.0368} &
\multirow{2}{*}{0.0842} & \multirow{2}{*}{0.0367} & \multirow{2}{*}{0.0846} \\
RNN & & & & & & &  & \\ \hline
DLinear & 
0.0543 & 0.0990 & 0.0551 & 0.0941 & 0.0573 & 0.0965 & 0.0556 & 0.0966\\ \hline
NLinear & 
0.0547 & 0.0991 & 0.0554 & 0.0940 & 0.0574 & 0.0961 & 0.0558 & 0.0964 \\ \hline
TiDE & 
0.0448 & 0.0885 & 0.0479 & 0.0908 & 0.0430 & 0.0912 & 0.0452 & 0.0902 \\ \hline
TFT & 
0.0375 & 0.0823 & 0.0434 & 0.0894 & 0.0395 & 0.0823 & 0.0401 & 0.0847 \\ \hline 
\cite{symeonidis2023wind} & 
0.0367 & 0.0810 & 0.0395 & \textbf{0.0830} & 0.0397 & 0.0819 & 0.0386 & 0.0820 \\ \hline \hline
Ours & 
\textbf{0.0331} & \textbf{0.0755} & 0.0388 & \textbf{0.0830} & \textbf{0.0366} & \textbf{0.0791} & \textbf{0.0362} & \textbf{0.0792}\\ \hline
\end{tabular}
}
\end{tabular}
\end{table}

\begin{figure}[h]
    \centering
    \includegraphics[width=\linewidth]{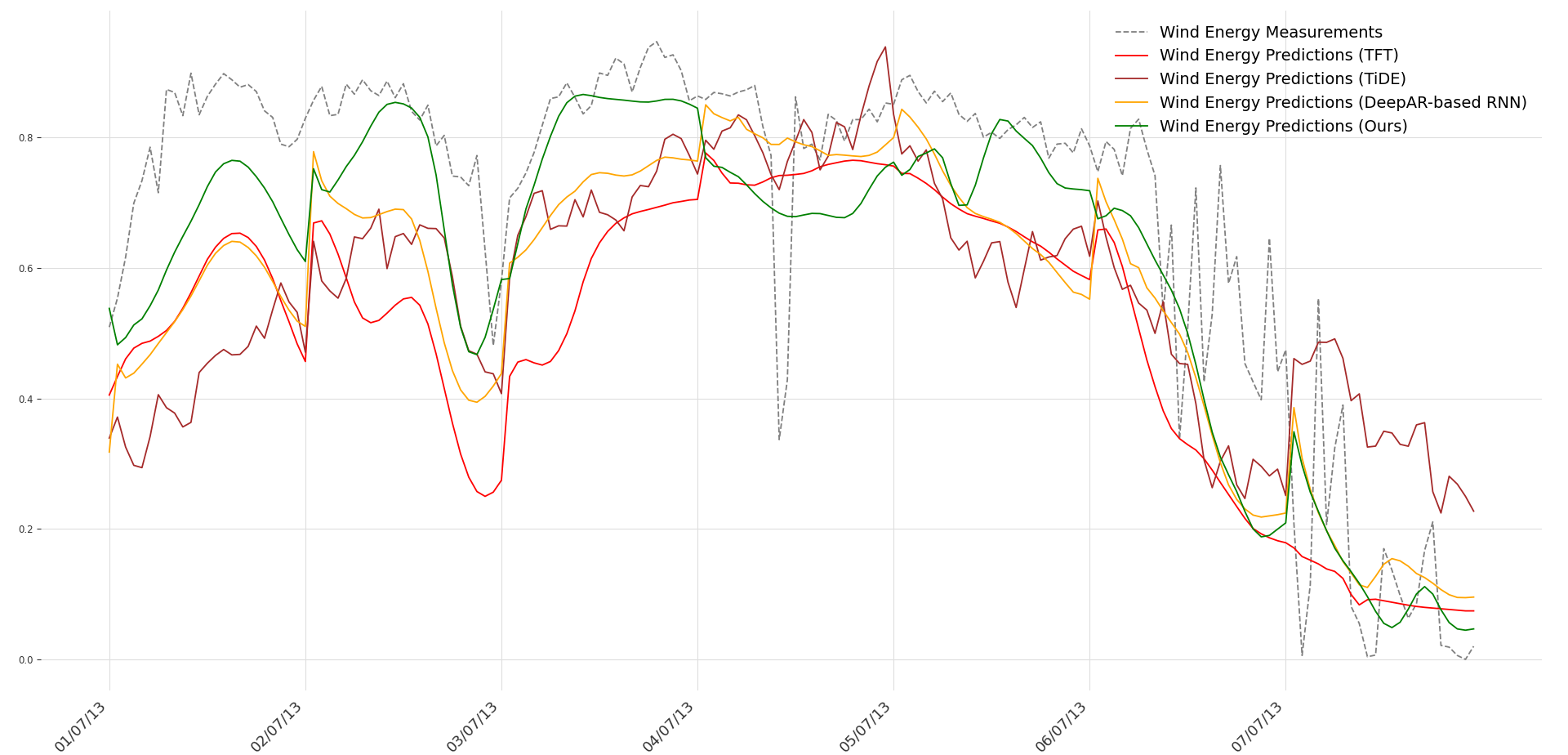}
    \caption{Day-ahead wind energy generation measurements (ground truth) and predictions for station VI of the AEMO-H dataset. Time period: 01-07/07/13.}
    \label{fig:ex_plot}
\end{figure}

\textbf{AEMO-H:} Tables~\ref{Tab:aemo_h_mae} and \ref{Tab:aemo_h_rmse} report the achieved MAE and RMSE for all employed methods in AEMO-H dataset. Our method attained the best results, for each of the twenty-two wind energy generation sites, on both evaluation metrics.  More specifically, it achieved MAE and RMSE of 0.098 and 0.138 respectively. TFT and the DeepAR-based RNN also achieved good results in terms of MAE and RMSE. In particular, the DeepAR-based RNN attained MAE of 0.089 in energy station VII, thereby aligning its performance with that of our proposed method. Figure~\ref{fig:ex_plot} depicts concatenated day-ahead energy predictions for seven days and station VI. The predictions were  generated by the best performing methods, namely DeepAR-based RNN, TFT, TiDE, and the proposed method. These predictions are compared against the actual energy generation measurements (ground truth). It is obvious that the predictions time-series generated by the proposed method follow in general, more faithfully, the ground truth. 

\clearpage
\begin{landscape}
\begin{table}[ht]
\centering
\resizebox{\columnwidth}{!}{
\begin{tabular}{|c|c|c|c|c|c|c|c|c|c|c|c|c|c|c|c|c|c|c|c|c|c|c||c|}
\hline
\multirow{2}{*}{\textbf{Method}} & \multicolumn{23}{c|}{\textbf{Energy Stations}} \\ \hhline{~-----------------------}

& \textbf{I} & \textbf{II} & \textbf{III} & \textbf{IV} & \textbf{V} & \textbf{VI}
& \textbf{VII} & \textbf{VIII} & \textbf{IX} & \textbf{X} & \textbf{XI}
& \textbf{XII} & \textbf{XIII} & \textbf{XIV} & \textbf{XV} & \textbf{XVI} 
& \textbf{XVII} & \textbf{XVIII} & \textbf{XIX} & \textbf{XX} & \textbf{XXI} 
& \textbf{XXII} & \textbf{All}  \\ \hline

TCNN &
0.217 & 0.201 & 0.243 & 0.207 & 0.204 & 0.231 & 0.227 & 0.232 & 0.234 & 0.226 & 0.233 & 0.236 & 0.235 & 0.245 & 0.249 & 0.195 & 0.203 & 0.245 & 0.206 & 0.264 & 0.217 & 0.237 & 0.227\\ \hline
N-HiTS &
0.168 & 0.171 & 0.184 & 0.155 & 0.171 & 0.188 & 0.170 & 0.180 & 0.189 & 0.183 & 0.199 & 0.190 & 0.182 & 0.199 & 0.204 & 0.149 & 0.162 & 0.192 & 0.171 & 0.210 & 0.184 & 0.193 & 0.182\\ \hline
Transformer &
0.162 & 0.146 & 0.172 & 0.141 & 0.147 & 0.176 & 0.154 & 0.157 & 0.164 & 0.161 & 0.169 & 0.174 & 0.168 & 0.177 & 0.182 & 0.138 & 0.151 & 0.171 & 0.152 & 0.180 & 0.164 & 0.174 & 0.163\\ \hline
Kalman &
\multirow{2}{*}{0.114} & \multirow{2}{*}{0.120} & \multirow{2}{*}{0.124} &
\multirow{2}{*}{0.102} & \multirow{2}{*}{0.117} & \multirow{2}{*}{0.156} &
\multirow{2}{*}{0.107} & \multirow{2}{*}{0.120} & \multirow{2}{*}{0.123} & 
\multirow{2}{*}{0.123} & \multirow{2}{*}{0.127} & \multirow{2}{*}{0.140} & 
\multirow{2}{*}{0.118} & \multirow{2}{*}{0.128} & \multirow{2}{*}{0.135} &
\multirow{2}{*}{0.102} & \multirow{2}{*}{0.117} & \multirow{2}{*}{0.122} &
\multirow{2}{*}{0.111} & \multirow{2}{*}{0.143} & \multirow{2}{*}{0.126} & 
\multirow{2}{*}{0.125} & \multirow{2}{*}{0.123} \\ 
Filtering & & & & & & & & & & & & & & & & & & & & & & & \\ \hline
Random &
\multirow{2}{*}{0.139} & \multirow{2}{*}{0.139} & \multirow{2}{*}{0.156} &
\multirow{2}{*}{0.128} & \multirow{2}{*}{0.140} & \multirow{2}{*}{0.169} &
\multirow{2}{*}{0.145} & \multirow{2}{*}{0.142} & \multirow{2}{*}{0.154} & 
\multirow{2}{*}{0.154} & \multirow{2}{*}{0.157} & \multirow{2}{*}{0.171} & 
\multirow{2}{*}{0.153} & \multirow{2}{*}{0.160} & \multirow{2}{*}{0.161} &
\multirow{2}{*}{0.122} & \multirow{2}{*}{0.139} & \multirow{2}{*}{0.159} &
\multirow{2}{*}{0.137} & \multirow{2}{*}{0.177} & \multirow{2}{*}{0.147} &
\multirow{2}{*}{0.154} & \multirow{2}{*}{0.150} \\ 
Forest & & & & & & & & & & & & & & & & & & & & & & & \\ \hline
DeepAR-based & 
\multirow{2}{*}{0.090} &
\multirow{2}{*}{0.095} &
\multirow{2}{*}{0.101} &
\multirow{2}{*}{0.080} & 
\multirow{2}{*}{0.104} &
\multirow{2}{*}{0.129} &
\multirow{2}{*}{\textbf{0.089}} &
\multirow{2}{*}{0.097} & 
\multirow{2}{*}{0.108} & 
\multirow{2}{*}{0.106} & 
\multirow{2}{*}{0.117} &
\multirow{2}{*}{0.124} &
\multirow{2}{*}{0.100} &
\multirow{2}{*}{0.111} &
\multirow{2}{*}{0.113} &
\multirow{2}{*}{0.085} &
\multirow{2}{*}{0.105} &
\multirow{2}{*}{0.107} &
\multirow{2}{*}{0.104} & 
\multirow{2}{*}{0.118} & 
\multirow{2}{*}{0.105} &
\multirow{2}{*}{0.099} & 
\multirow{2}{*}{0.104} \\ 
RNN & 
 & & & & & & & & & & & & & & & & & & & & & & \\ \hline 
DLinear &
0.155 & 0.159 & 0.180 & 0.136 & 0.157 & 0.179 & 0.156 & 0.157 & 0.168 & 0.164 & 0.178 & 0.189 & 0.171 & 0.185 & 0.193 & 0.134 & 0.148 & 0.178 & 0.155 & 0.197 & 0.169 & 0.177 & 0.168\\ \hline
NLinear &
0.165 & 0.174 & 0.186 & 0.145 & 0.160 & 0.179 & 0.162 & 0.168 & 0.174 & 0.182 & 0.196 & 0.191 & 0.177 & 0.203 & 0.211 & 0.144 & 0.155 & 0.197 & 0.161 & 0.204 & 0.186 & 0.195 & 0.178\\ \hline
TiDE &
0.111 & 0.123 & 0.127 & 0.096 & 0.116 & 0.146 & 0.110 & 0.112 & 0.126 & 0.133 & 0.130 & 0.149 & 0.123 & 0.136 & 0.139 & 0.100 & 0.125 & 0.133 & 0.128 & 0.148 & 0.126 & 0.123 & 0.125\\ \hline
TFT &
0.100 & 0.101 & 0.105 & 0.082 & 0.106 & 0.141 & 0.092 & 0.099 & 0.119 & 0.114 & 0.115 & 0.129 & 0.102 & 0.114 & 0.112 & 0.089 & 0.110 & 0.114 & 0.105 & 0.123 & 0.107 & 0.104 & 0.108\\ \hline 
\cite{symeonidis2023wind} &
0.125 & 0.128 & 0.136 & 0.112 & 0.125 & 0.141 & 0.127 & 0.125 & 0.139 & 0.130 & 0.144 & 0.152 & 0.136 & 0.141 & 0.144 & 0.105 & 0.120 & 0.142 & 0.128 & 0.157 & 0.134 & 0.133 & 0.133 \\ \hline \hline
Ours &
\textbf{0.082} & \textbf{0.093} & \textbf{0.096} & \textbf{0.079} & \textbf{0.096} & \textbf{0.105} & \textbf{0.089} & \textbf{0.092} & \textbf{0.105} & \textbf{0.102} & \textbf{0.110} & \textbf{0.116} & \textbf{0.095} & \textbf{0.108} & \textbf{0.107} & \textbf{0.077} & \textbf{0.093} & \textbf{0.105} & \textbf{0.099} & \textbf{0.115} & \textbf{0.098} & \textbf{0.093} & \textbf{0.098} \\ \hline
\end{tabular}
}
\caption{Mean Absolute Error (MAE) of the compared methods in AEMO-H dataset.}
\label{Tab:aemo_h_mae}
\end{table}

\begin{table}[ht]
\centering
\resizebox{\columnwidth}{!}{
\begin{tabular}{|c|c|c|c|c|c|c|c|c|c|c|c|c|c|c|c|c|c|c|c|c|c|c||c|}
\hline
\multirow{2}{*}{\textbf{Method}} & \multicolumn{23}{c|}{\textbf{Energy Stations}} \\ \hhline{~-----------------------}

& \textbf{I} & \textbf{II} & \textbf{III} & \textbf{IV} & \textbf{V} & \textbf{VI}
& \textbf{VII} & \textbf{VIII} & \textbf{IX} & \textbf{X} & \textbf{XI}
& \textbf{XII} & \textbf{XIII} & \textbf{XIV} & \textbf{XV} & \textbf{XVI} 
& \textbf{XVII} & \textbf{XVIII} & \textbf{XIX} & \textbf{XX} & \textbf{XXI} 
& \textbf{XXII} & \textbf{All}  \\ \hline

TCNN &
0.273 & 0.241 & 0.294 & 0.264 & 0.247 & 0.272 & 0.275 & 0.285 & 0.280 & 0.277 & 0.290 & 0.279 & 0.284 & 0.295 & 0.300 & 0.243 & 0.255 & 0.294 & 0.256 & 0.307 & 0.270 & 0.293 & 0.276\\ \hline
N-HiTS &
0.230 & 0.217 & 0.238 & 0.209 & 0.221 & 0.240 & 0.219 & 0.231 & 0.240 & 0.237 & 0.255 & 0.241 & 0.231 & 0.255 & 0.264 & 0.198 & 0.215 & 0.246 & 0.224 & 0.265 & 0.239 & 0.253 & 0.235\\ \hline
Transformer &
0.218 & 0.189 & 0.220 & 0.192 & 0.195 & 0.226 & 0.203 & 0.209 & 0.212 & 0.211 & 0.224 & 0.219 & 0.215 & 0.227 & 0.236 & 0.182 & 0.198 & 0.221 & 0.206 & 0.233 & 0.215 & 0.226 & 0.213\\ \hline
Kalman &
\multirow{2}{*}{0.163} & \multirow{2}{*}{0.160} & \multirow{2}{*}{0.162} &
\multirow{2}{*}{0.138} & \multirow{2}{*}{0.160} & \multirow{2}{*}{0.205} &
\multirow{2}{*}{0.137} & \multirow{2}{*}{0.161} & \multirow{2}{*}{0.161} & 
\multirow{2}{*}{0.161} & \multirow{2}{*}{0.170} & \multirow{2}{*}{0.179} & 
\multirow{2}{*}{0.152} & \multirow{2}{*}{0.166} & \multirow{2}{*}{0.178} &
\multirow{2}{*}{0.136} & \multirow{2}{*}{0.163} & \multirow{2}{*}{0.160} &
\multirow{2}{*}{0.157} & \multirow{2}{*}{0.185} & \multirow{2}{*}{0.172} & 
\multirow{2}{*}{0.171} & \multirow{2}{*}{0.164} \\ 
Filtering & & & & & & & & & & & & & & & & & & & & & & & \\ \hline
Random &
\multirow{2}{*}{0.188} & \multirow{2}{*}{0.179} & \multirow{2}{*}{0.201} &
\multirow{2}{*}{0.174} & \multirow{2}{*}{0.187} & \multirow{2}{*}{0.220} &
\multirow{2}{*}{0.186} & \multirow{2}{*}{0.185} & \multirow{2}{*}{0.201} & 
\multirow{2}{*}{0.200} & \multirow{2}{*}{0.207} & \multirow{2}{*}{0.212} & 
\multirow{2}{*}{0.196} & \multirow{2}{*}{0.203} & \multirow{2}{*}{0.207} &
\multirow{2}{*}{0.163} & \multirow{2}{*}{0.186} & \multirow{2}{*}{0.201} &
\multirow{2}{*}{0.185} & \multirow{2}{*}{0.223} & \multirow{2}{*}{0.195} &
\multirow{2}{*}{0.206} & \multirow{2}{*}{0.196} \\ 
Forest & & & & & & & & & & & & & & & & & & & & & & & \\ \hline
DeepAR-based & 
\multirow{2}{*}{0.136} &
\multirow{2}{*}{0.137} &
\multirow{2}{*}{0.142} & 
\multirow{2}{*}{0.116} & 
\multirow{2}{*}{0.149} & 
\multirow{2}{*}{0.177} & 
\multirow{2}{*}{0.120} & 
\multirow{2}{*}{0.136} & 
\multirow{2}{*}{0.148} & 
\multirow{2}{*}{0.148} & 
\multirow{2}{*}{0.163} & 
\multirow{2}{*}{0.168} & 
\multirow{2}{*}{0.134} & 
\multirow{2}{*}{0.153} & 
\multirow{2}{*}{0.161} &
\multirow{2}{*}{0.120} &
\multirow{2}{*}{0.156} & 
\multirow{2}{*}{0.151} &
\multirow{2}{*}{0.153} &
\multirow{2}{*}{0.166} & 
\multirow{2}{*}{0.156} & 
\multirow{2}{*}{0.143} & 
\multirow{2}{*}{0.147} \\
RNN & 
 & & & & & & & & & & & & & & & & & & & & & & \\ \hline 
DLinear &
0.212 & 0.205 & 0.235 & 0.188 & 0.209 & 0.233 & 0.205 & 0.207 & 0.220 & 0.217 & 0.236 & 0.242 & 0.222 & 0.240 & 0.252 & 0.181 & 0.202 & 0.233 & 0.208 & 0.256 & 0.226 & 0.235 & 0.221\\ \hline
NLinear &
0.221 & 0.219 & 0.238 & 0.196 & 0.210 & 0.230 & 0.208 & 0.217 & 0.225 & 0.233 & 0.251 & 0.241 & 0.226 & 0.256 & 0.266 & 0.191 & 0.208 & 0.248 & 0.214 & 0.260 & 0.241 & 0.253 & 0.230 \\ \hline
TiDE &
0.160 & 0.165 & 0.171 & 0.134 & 0.159 & 0.191 & 0.142 & 0.152 & 0.169 & 0.177 & 0.180 & 0.196 & 0.163 & 0.181 & 0.189 & 0.136 & 0.175 & 0.177 & 0.177 & 0.198 & 0.178 & 0.172 & 0.170 \\ \hline
TFT &
0.148 & 0.140 & 0.146 & 0.117 & 0.152 & 0.186 & 0.122 & 0.138 & 0.166 & 0.153 & 0.163 & 0.171 & 0.137 & 0.156 & 0.160 & 0.123 & 0.157 & 0.155 & 0.155 & 0.171 & 0.159 & 0.146 & 0.151\\ \hline 
\cite{symeonidis2023wind} &
0.179 & 0.172 & 0.185 & 0.155 & 0.173 & 0.189 & 0.165 & 0.168 & 0.185 & 0.176 & 0.194 & 0.199 & 0.182 & 0.188 & 0.194 & 0.145 & 0.170 & 0.188 & 0.175 & 0.208 & 0.180 & 0.188 & 0.180 \\ \hline \hline
Ours &
\textbf{0.124} & \textbf{0.130} & \textbf{0.133} & \textbf{0.113} & \textbf{0.137} & \textbf{0.146} & \textbf{0.119} & \textbf{0.129} & \textbf{0.145} & \textbf{0.142} & \textbf{0.155} & \textbf{0.159} & \textbf{0.129} & \textbf{0.148} & \textbf{0.152} & \textbf{0.110} & \textbf{0.141} & \textbf{0.145} & \textbf{0.146} & \textbf{0.160} & \textbf{0.141} & \textbf{0.134} & \textbf{0.138} \\ \hline
\end{tabular}
}
\caption{Root Mean Square Error (RMSE) of the compared methods in AEMO-H dataset.}
\label{Tab:aemo_h_rmse}
\end{table}
\end{landscape}
\clearpage

\textbf{ENTSO-E wind/solar:}  The MAE and RMSE values achieved by  the proposed and the baseline methods in ENTSO-E wind and solar datasets are reported in Table~\ref{Tab:entsoe_mae_rmse}. The proposed method attained the best results, for both evaluation metrics. In ENTSO-E wind, it achieved top MAE and RMSE values in normalized data, namely 0.065 and 0.087 respectively. TiDE and the method in \cite{symeonidis2023wind} achieved also good results in both metrics. Similarly in ENTSO-E solar, our proposed method achieved the best performance  (0.035 MAE and 0.072 RMSE)  in normalized data. Again, TiDE and the method proposed in \cite{symeonidis2023wind} attained favourable results.

\begin{table}[ht]
\centering
\resizebox{\columnwidth}{!}{
\begin{tabular}{|c|cc|cc|}
\hline
\multirow{3}{*}{\textbf{Method}} & \multicolumn{2}{c|}{\textbf{Wind}} & \multicolumn{2}{c|}{\textbf{Solar}}  \\ 

 & \textbf{MAE} & \textbf{RMSE} & \textbf{MAE} & \textbf{RMSE}  \\
  & \textbf{(nrm. / MW)} & \textbf{(nrm. / MW)} & \textbf{(nrm. / MW)} & \textbf{(nrm. / MW)}  \\ \hline
TCNN &  0.203 / 532.60 & 0.251 / 660.21 & 0.063 / 134.15 & 0.123 / 260.94 \\ \hline
N-HiTS & 0.141 / 370.53 & 0.190 / 498.62 & 0.060 / 126.67 & 0.112 / 237.44 \\ \hline
Transformer & 0.135 / 354.47 & 0.180 / 474.12 & 0.058 / 123.36 & 0.111 / 235.31 \\ \hline
Kalman & \multirow{2}{*}{0.108 / 284.37} & \multirow{2}{*}{0.155 / 408.19} &
\multirow{2}{*}{0.143 / 301.87} & \multirow{2}{*}{0.237 / 502.24} \\
Filtering & & & & \\ \hline
Random & \multirow{2}{*}{0.116 / 305.34} & \multirow{2}{*}{0.157 / 413.61} &
\multirow{2}{*}{0.114 / 240.74} & \multirow{2}{*}{0.209 / 442.07} \\
Forest & & & & \\ \hline
DeepAR-based & \multirow{2}{*}{0.093 / 245.81} & \multirow{2}{*}{0.132 / 346.13} & 
\multirow{2}{*}{0.095 / 201.10} & 
\multirow{2}{*}{0.175 / 370.48} \\
RNN & & & & \\ \hline

DLinear & 0.107 / 281.29 & 0.145 / 381.33 & 0.062 / 130.47 & 0.105 / 222.74 \\ \hline
NLinear & 0.110 / 289.70 & 0.147 / 387.54 & 0.061 / 128.10 & 0.104 / 219.45 \\ \hline
TiDE & 0.078 / 206.14 & 0.105 / 275.28 & 0.040 / 83.83 & 0.078 / 164.49 \\ \hline
TFT & 0.084 / 221.25 & 0.116 / 304.29 & 0.063 / 134.15 & 0.123 / 260.94 \\ \hline 
\cite{symeonidis2023wind} &  0.076 / 200.71  & 0.102 / 267.83 &  0.040 / 84.64 & 0.080 / 168.53 \\ \hline \hline
Ours &  \textbf{0.065} / \textbf{170.37}  & \textbf{0.087} / \textbf{229.04} &  \textbf{0.035} / \textbf{73.70} & \textbf{0.072} / \textbf{151.47} \\ \hline

\end{tabular}
}
\caption{Mean Absolute Error (MAE) and Root Mean Square Error (RMSE) of the compared methods in ENTSO-E wind/solar dataset, for both normalized and raw data.}
\label{Tab:entsoe_mae_rmse}
\end{table}

\subsection{Discussion}

Overall, the proposed wind and solar energy forecasting method achieved the best performance in terms of average - across the energy generation stations- MAE and RMSE. This superiority with respect to the competing methods was manifested in all five datasets. In addition, when measuring the performance  on each energy generation site separately, the proposed method achieved top results on most of the cases. The obtained results highlight the ability of the method to generate enriched latent representations for energy generation data and weather data by exploring spatio-temporal patterns among them. It shall be also noted that  the temporal processing of weather data in the proposed method, facilitated by the UTCAE, consistently demonstrates enhanced performance  with respects to our  preliminary approach presented  in \cite{symeonidis2023wind}.

\begin{figure}[ht]
    \centering
    \includegraphics[width=0.9\linewidth]{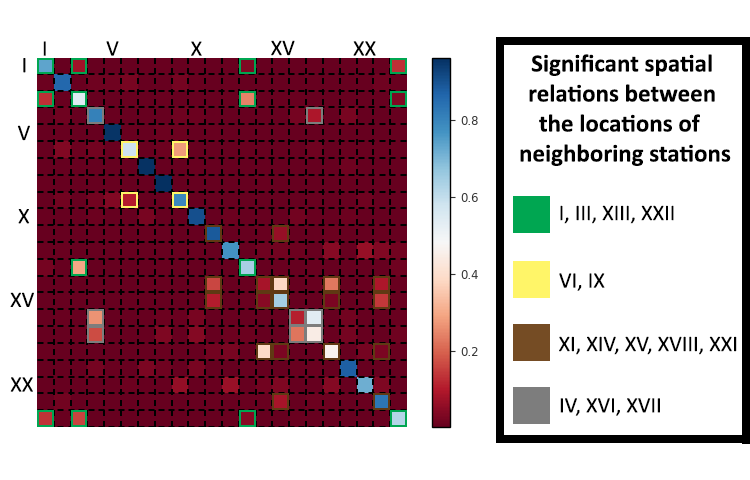}
    \caption{Spatial attention weights matrix, computed by the proposed method for the AEMO-H dataset. The locations of the energy stations, represented as rows, coincide with the weather data locations, represented as columns.}
    \label{fig:sp_att_m}
\end{figure}

\begin{figure}[H]
    \centering
    \includegraphics[width=0.9\linewidth]{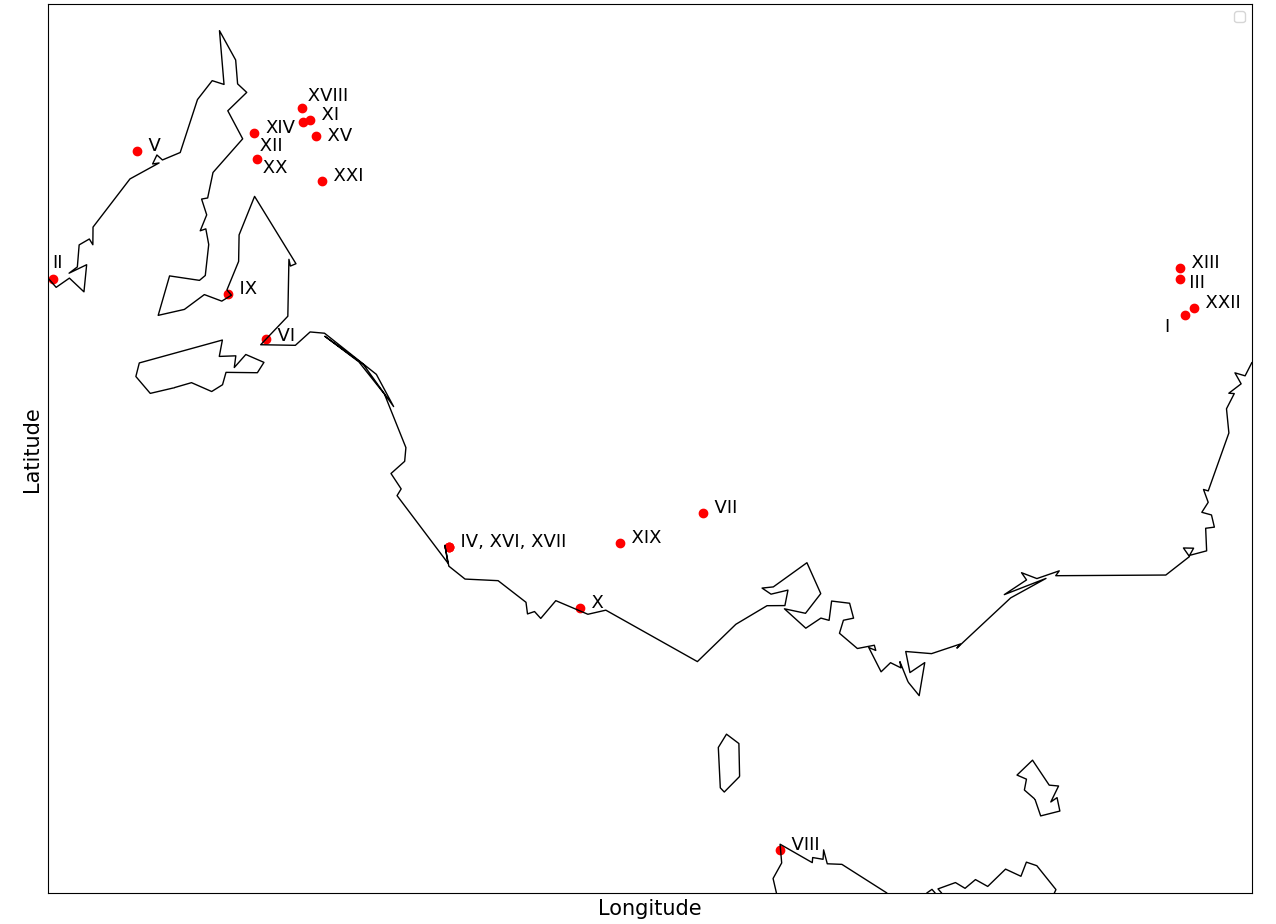}
    \caption{Locations of the 22 energy generation stations in AEMO-H (south-east Australia). The weather data locations coincide with the locations of the stations.}
    \label{fig:aemo_map}
\end{figure}

Aiming to demonstrate the efficacy of the proposed methodology in establishing meaningful spatial relationships between the locations of energy generation stations and the locations of weather data, we compared for the AEMO-H dataset the spatial attention weights matrix  illustrated in Figure~\ref{fig:sp_att_m} with the corresponding geographical locations  depicted in Figure~\ref{fig:aemo_map}. Indeed, the method forms significant spatial relations between neighboring stations such as stations a) I, III, XIII, XXII b) VI, IX, c) XI, XIV, XV, XVIII, XXI and d) IV, XVI, XVII. To aid the reader, the attention weights corresponding to these stations are marked in the figure with outlines of different colors.

Moreover, in order to thoroughly evaluate the importance of spatial relations, we conducted a comprehensive examination of the proposed methodology across three distinct scenarios, using data from the AEMO-H dataset. In all these scenarios, the proposed method is applied  for day-ahead energy generation forecasting at eleven stations, namely stations IV, XIII-XXII, of this datasets. The first scenario aimed at analysing the method's performance in the case where weather data for the locations  of the energy stations are available. Thus, in this scenario, weather data for the locations  of stations IV, XII-XXII are provided. The competing methods, namely Kalman Filtering, DeepAR-based RNN, TiDE and TFT, were also applied under this scenario.
 The second scenario was designed so as to assess the proposed method's ability to discover spatial relations when the locations of the weather data don't align with these of the energy stations. Thus, in this scenario, weather data for locations corresponding to the positions of stations I-III and V-XII (namely stations that are not included in the experiment) are provided.
In the third scenario, the examination involves  weather data for the locations  of the energy stations, as well as supplementary weather data, for additional locations. Thus, weather data for locations corresponding to the  positions of all AEMO-H stations, namely stations I-XXII are provided.

As illustrated in Table~\ref{Tab:aemo_extra}, the proposed method attains top MAE (0.099) and RMSE (0.142) when  weather data are available for the  positions of the energy stations. A similar performance was also achieved when utilizing weather data corresponding to the positions of the aforementioned stations, as well as from locations corresponding to stations not involved in the associated evaluation (scenario 3). In the case where weather data were provided  for locations which do not coincide with the positions of the energy stations (scenario 2), the proposed method yielded MAE and RMSE of 0.109 and 0.152, respectively. Thus, in this challenging case, the method managed to attain a decent performance, being third behind the DeepAR-based RNN and TFT, which, as mentioned above, utilized weather data from the locations of the stations. These findings highlight the significance of  weather data, especially when their locations  coincide with the positions of the energy stations. Results also show that the proposed method can  perform well when provided with weather data for locations in close proximity to the energy stations.

\begin{table}[ht]
\centering
\begin{tabular}{|c|c|c|c|}
\hline
\textbf{Method} & \textbf{Weather Data for Specified Stations} & \textbf{MAE} & \textbf{RMSE}  \\ \hline
Kalman Filtering & IV, XII-XXII (scenario 1) & 0.121  & 0.162 \\ \hline
DeepAR-based RNN & IV, XII-XXII (scenario 1)& 0.102 & 0.146 \\ \hline
TiDE & IV, XII-XXII (scenario 1)& 0.125 & 0.171  \\ \hline
TFT & IV, XII-XXII (scenario 1)& 0.106 & 0.149 \\ \hline \hline
\multirow{3}{*}{Ours} & IV, XII-XXII (scenario 1) & \textbf{0.099} & 0.142  \\ \cmidrule(){2-4}
&  I-III, V-XII (scenario 2)& 0.109 & 0.152 \\  \cmidrule{2-4}
& I-XXII (scenario 3)& \textbf{0.099} & \textbf{0.141}  \\ \hline
\end{tabular}

\caption{Average Mean Absolute Error (MAE) and average Root Mean Square Error (RMSE) for the top-five compared methods,  in  day-ahead energy forecasting for stations IV and XII-XXII of the AEMO-H dataset. The proposed methodology undergoes evaluation in three distinct scenaria, considering the spatial availability of the provided weather data (second column).}
\label{Tab:aemo_extra}

\label{discussion}
\end{table}
\textbf{}

\section{Conclusions}\label{conclusion}

The efficient integration of Renewable Energy Sources (RES) into existing power systems  highly depends on the accuracy of energy generation forecasting. This paper proposes a novel, deterministic multi-site and multi-step (i.e., for multiple time instances) wind and solar energy generation forecasting methodology, which utilizes past-time energy generation measurements as well as past- and future-time weather data from multiple locations. The  weather data locations and the locations of the energy generation stations may not coincide with each other and no information, such as geolocation coordinates, needs to be  provided. The proposed method utilizes a U-shaped Temporal Convolutional Auto-Encoder architecture to process time-series data related to weather and energy generation across each site. Furthermore, inspired by the multi-head scaled-dot product attention mechanism, we introduce the Multi-sized Kernels convolutional Spatio-Temporal Attention, in order to effectively transfer temporal patterns from weather data to energy data. The results of the conducted experimental evaluation, in a day-ahead forecasting scenario, on three wind and two solar energy generation forecasting datasets, highlight the potential of the proposed method and its superiority against eleven state-of-the-art time-series forecasting methods. Future extensions will focus on validating the effectiveness of the method on various forecasting horizons, temporal resolutions, and datasets where the locations of the weather data don't correspond with the locations of the energy stations.

\backmatter

\bmhead{Acknowledgments}

This work is co-financed by the European Regional Development Fund of the European Union and Greek national funds through the Operational Program Competitiveness, Entrepreneurship and Innovation, under the call RESEARCH - CREATE - INNOVATE (project code: T2EDK-03048).

\section*{Declarations}



\bmhead{Code availability} The code used for the conducted experiments can be found at: \url{https://github.com/charsyme/Efficient-Deterministic-Renewable-Energy-Forecasting-Guided-by-Multiple-Location-Weather-Data}.

\bmhead{Data availability} The following datasets are available:

\noindent GEFCom2014-wind and GEFCom2014-solar datasets are available, at \url{https://dx.doi.org/10.1016/j.ijforecast.2016.02.001}.

\bigbreak

\noindent AEMO dataset is publicly available at \url{https://pureportal.strath.ac.uk/en/publications/very-short-term-probabilistic-wind-power-forecasts-by-sparse-vect}. The weather data necessary for reproducing the AEMO-H dataset can be obtained from the the Solcast online service at \url{https://solcast.com/}, utilizing the geolocation coordinates of the stations featured in the AEMO dataset.

\bigbreak

\noindent The energy generation data of ENTSO-E wind/solar datasets can be obtained from the European Network of Transmission System Operators for Electricity at \url{https://transparency.entsoe.eu}. The weather data necessary for reproducing the aforementioned datasets can be obtained from the OpenWeather online service at \url{https://openweathermap.org/}, utilizing the geolocation coordinates of the 29 locations.








\bibliography{sn-bibliography}

\end{document}